\pdfoutput=1

\documentclass[11pt]{article}

\usepackage[preprint]{acl}

\usepackage{times}
\usepackage{latexsym}

\usepackage[T1]{fontenc}

\usepackage[utf8]{inputenc}

\usepackage{microtype}

\usepackage{inconsolata}

\usepackage{forest}

\usepackage{amsmath}
\usepackage{graphicx}
\usepackage{dsfont}
\usepackage{caption}
\usepackage{tabularx}
\usepackage{wrapfig}
\usepackage{multirow}
\usepackage{colortbl}  
\usepackage{listings}
\usepackage{collcell}
\usepackage{amssymb}
\usepackage{array}
\usepackage{enumitem}
\usepackage{subcaption}
\usepackage{arydshln}   
\usepackage{graphicx}   
\usepackage{array}      
\usepackage{booktabs}  

\title{A Survey of Useful LLM Evaluation}

\author{
    Ji-Lun Peng$^*$\quad
    Sijia Cheng$^*$\quad
    Egil Diau$^*$\quad
    Yung-Yu Shih$^*$\quad
    \\
    \textbf{Po-Heng Chen$^*$\quad Yen-Ting Lin\quad
    Yun-Nung Chen}
    \\
  National Taiwan University, Taipei, Taiwan \\
  \texttt{\{b09207002, r11922184, r12922a03, r12944007, r11922044\}@ntu.edu.tw}\\
  \texttt{\{ytl, y.v.chen\}ieee.org}
}

\begin{document}
\maketitle
\begin{abstract}
LLMs have gotten attention across various research domains due to their exceptional performance on a wide range of complex tasks. Therefore, refined methods to evaluate the capabilities of LLMs are needed to determine the tasks and responsibility they should undertake. Our study mainly discussed how LLMs, as useful tools, should be effectively assessed. We proposed the two-stage framework: from ``core ability'' to ``agent'', clearly explaining how LLMs can be applied based on their specific capabilities, along with the evaluation methods in each stage. Core ability refers to the capabilities that LLMs need in order to generate high-quality natural language texts. After confirming LLMs possess core ability, they can solve real-world and complex tasks as agent. In the "core ability" stage, we discussed the reasoning ability, societal impact, and domain knowledge of LLMs. In the ``agent'' stage, we demonstrated embodied action, planning, and tool learning of LLMs agent applications. Finally, we examined the challenges currently confronting the evaluation methods for LLMs, as well as the directions for future development.\footnote{\url{https://github.com/MiuLab/EvalLLM-Survey}}
\begingroup\def\thefootnote{\rm *}\footnotetext{Equal contribution.}\endgroup
\end{abstract}

\section{Introduction}
\begin{figure*}[hbt!]
\centering
\includegraphics[width=\linewidth]{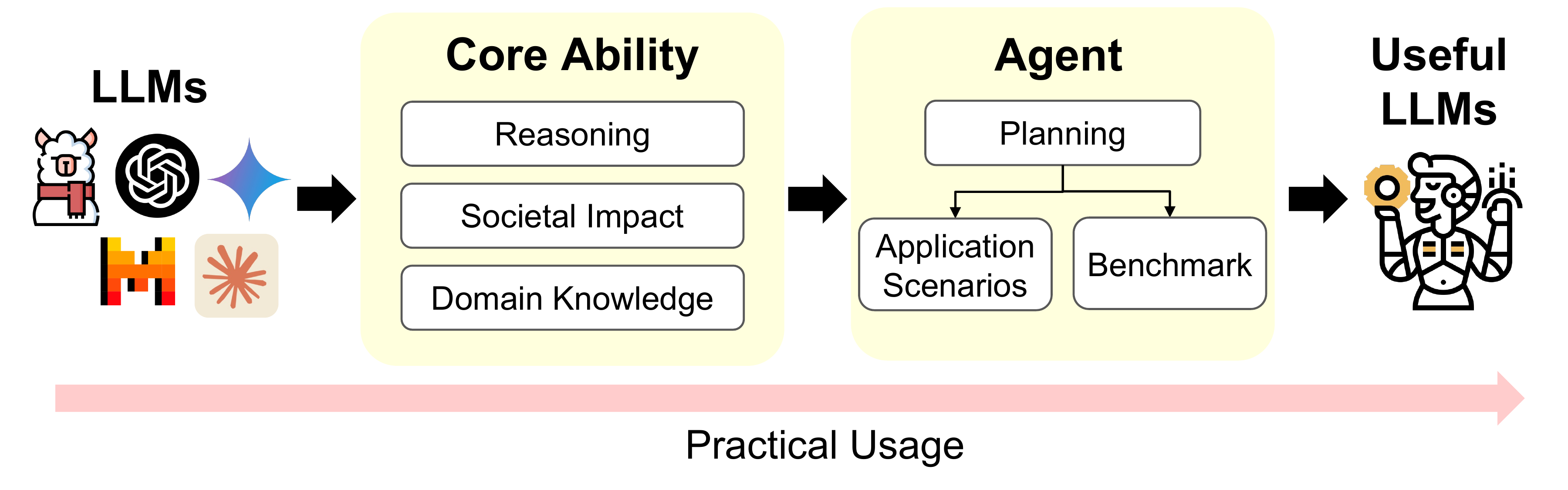}
\caption{The two-stage framework of our LLMs evaluation.}
\label{fig:framework}
\end{figure*}
\subsection{Artificial Intelligence and Large Language Model}
Artificial intelligence (AI) simulates human behavior to complete multiple tasks needing human intelligence. The first models of AI tried to simulate the function of a single neuron with feedforward, simple input-output functions \cite{muthukrishnan2020brief}. As time has progressed, a variety of machine learning (ML) and deep learning (DL) models have been developed. They are not only capable of identifying patterns from vast amounts of data, but they can also make predictions, and even handle unstructured data such as text, images, and audio.  Recently, the Transformer architecture \cite{vaswani2017attention} was proposed,  allowing word embeddings to be context-dependent, and model training to be scaled up \cite{min2023recent}. Therefore, researchers gradually increased the parameters in pre-trained language models trying to reach better performance. Using the Generative Pre-trained Transformer (GPT) series as an illustration, the progression in complexity and models' capability is marked by a significant increase in the number of parameters: GPT-1 \cite{radford2018improving} has 117 million parameters, GPT-2 \cite{radford2019language} expands this to 1.5 billion parameters, and GPT-3 \cite{mann2020language} further escalates to 175 billion parameters. Moreover, GPT-4 released by OpenAI with much larger model size could accept image and text inputs and produce text outputs, and exhibited human-level performance on various professional and academic benchmarks \cite{achiam2023gpt}. The models mentioned above, due to their tremendous size, are referred to as LLMs. They have gotten attention across various research domains due to their exceptional performance on a wide range of complex tasks.

\subsection{Why Evaluating LLMs is Important}
The early works testing model’s intelligence referred to as the Turing Test, raising the question of whether machines could imitate human intelligence and made people fail to differentiate \cite{pinar2000turing}. Evaluating AI is vital as it helps us gauge the real-world capabilities and limitations of AI systems. As AI technologies improve, particularly in areas like software testing and structural engineering, they can sometimes perform better than humans. However, we need clear benchmarks to make sure these technologies are both reliable and effective \cite{salehi2018emerging}. With the rapid evolution of LLMs, refined methods to evaluate the capabilities of LLMs is needed to determine the tasks and responsibility they should undertake. Because LLMs exhibit a broad spectrum of capabilities beyond the specific task they are trained for: predicting the next words of human-written texts \cite{nolfi2023unexpected}, such as formal linguistic competence \cite{mahowald2023dissociating}, factual knowledge \cite{petroni2019language}, and even theory of mind skills \cite{kosinski2023theory}, we should design benchmarks or evaluation methods specific to each task or domain. In current benchmarks, the comprehensive abilities of LLMs are automatically evaluated through tasks spanning multiple domains such as HELM \cite{liang2022holistic} and BIG-Bench \cite{srivastava2022beyond}, or by generating human feedback automatically like AlpacaFarm \cite{dubois2024alpacafarm} and MT-bench \cite{zheng2024judging}. However, when LLMs are required to perform specific tasks, the existence of evaluation methods tailored to those tasks becomes potential. This allows for a comparison of different models' capabilities under identical tasks to select the best performer. In this study, we categorizes LLMs' distinct abilities, systematically reviews existing evaluation methods under each category, and discusses how LLMs, as "useful" tools, should be effectively assessed.

\subsection{The Roadmap of Useful LLMs}
To determine whether LLMs are capable to become useful tools, we should split LLMs' capabilities into "core ability" and "agent", and discuss them respectively. Core ability refers to the capabilities that LLMs need in order to generate high-quality natural language texts, which are the foundation of performing complex behaviors.

Firstly, LLMs must possess the capability for reasoning, as during interactions with humans, they are required to deduce arguments step by step to engage in effective discussion. Furthermore, the societal impact of LLMs needs significant attention, for LLMs must be perceived as safe and trustworthy for humans to believe in and actively use them. Lastly, LLMs should have knowledge across various domains, and they can assist humans in solving problems occurring withing diverse fields.

Upon confirming that LLMs possess these core abilities, we can utilize LLMs to perform complex behaviors to deal with real-world problems, which we define as agent. For instance, LLMs agents can perform planning, generating an explicit deliberation process that chooses and organizes actions by anticipating their expected outcomes \cite{ghallab2004automated}. Then, LLMs agent can solve tasks in various scenarios such as using tools, creating tools, navigating embodied robots, and so on.

Even though LLMs can display the aforementioned capabilities, comprehensive evaluation methods are necessary to ensure that LLMs achieve as satisfactory level of performance in executing each task. Existing papers on LLM evaluation methods, including \citet{guo2023evaluating} and \citet{chang2023survey}, provide a thorough review of evaluation approaches for various aspects of LLMs, yet no study has offered a phased framework to explore the usability of LLMs. Hence, this paper proposes a two-stage framework to examine whether LLMs are sufficiently useful tools (\autoref{fig:framework}).

\subsection{Study Overview}
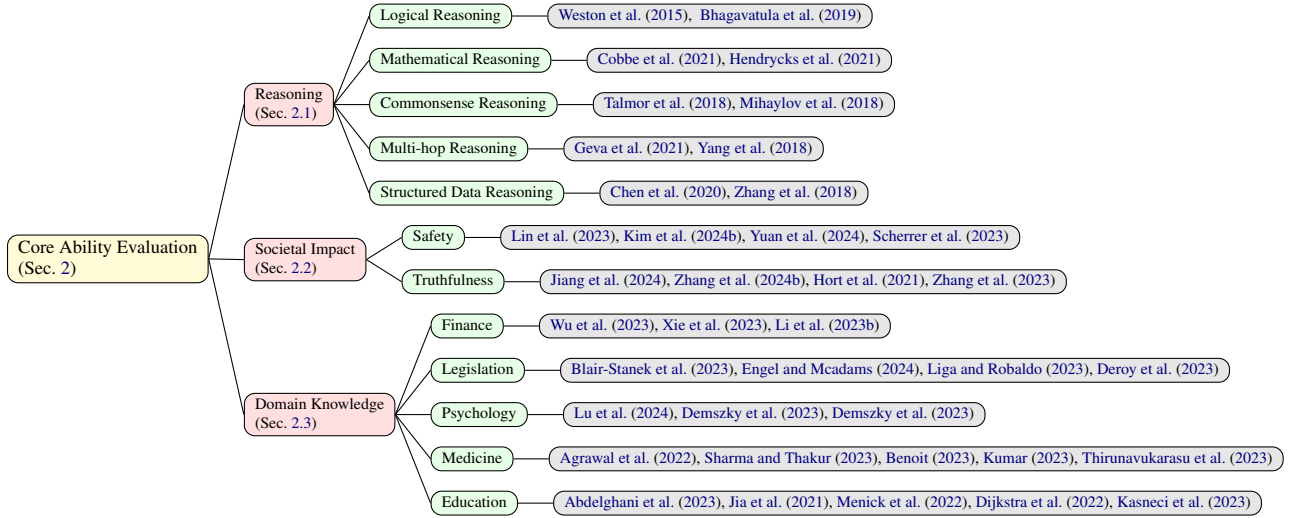
\begin{figure*}
\centering
\begin{forest}
  for tree={
  grow=east,
  reversed=true,
  anchor=base west,
  parent anchor=east,
  child anchor=west,
  base=left,
  font=\tiny,
  rectangle,
  draw,
  rounded corners,align=left,
  inner xsep=4pt,
  inner ysep=1pt,
  },
  where level=1{font=\tiny,fill=pink!50}{},
  where level=2{font=\tiny,fill=green!10}{},
  where level=3{font=\tiny,fill=gray!20}{},
    [Core Ability Evaluation\\(Sec.~\ref{sec:language_evaluation}),fill=yellow!20,font=\scriptsize %
        [Reasoning\\(Sec.~\ref{sec:reasoning})
          [Logical Reasoning
            [\citet{weston2015towards}\text{, } \citet{bhagavatula2019abductive}]
          ]
          [Mathematical Reasoning
            [\citet{cobbe2021training}\text{, }\citet{hendrycks2021measuring}]
          ]
          [Commonsense Reasoning
            [\citet{talmor2018commonsenseqa}\text{, }\citet{OpenBookQA2018}]
          ]
          [Multi-hop Reasoning
            [\citet{geva2021did}\text{, }\citet{yang2018hotpotqa}]
          ]
          [Structured Data Reasoning
            [\citet{chen2020hybridqa}\text{, }\citet{zhang2017variational}]
          ]
        ]
        [Societal Impact\\(Sec.~\ref{sec:societal_impact}) 
            [Safety[\citet{lin2023toxicchat}\text{, }\citet{kim2024propile}\text{, }\citet{yuan2024rigorllm}\text{, }\citet{scherrer2023evaluating}]
          ]
            [Truthfulness
            [\citet{jiang2024hal}\text{, }\citet{zhang2024self}\text{, }\citet{hort2021fairea}\text{, }\citet{Zhang_2023}]
          ]
        ]
        [Domain Knowledge\\(Sec.~\ref{sec:domain_knowledge})
            [Finance
             [\citet{wu2023bloomberggpt}\text{, }\citet{xie2023pixiu}\text{, }\citet{li2023large}]
            ]
            [Legislation
             [\citet{blair2023can}\text{, }\citet{engel2024asking}\text{, }\citet{liga2023fine}\text{, }\citet{deroy2023ready}
             ]
            ]
            [Psychology
             [\citet{lu2024gpt}\text{, }\citet{demszky2023using}\text{, }\citet{demszky2023using}
             ]
            ]
            [Medicine
             [\citet{agrawal2022large}\text{, }\citet{sharma2023chatgpt}\text{, }\citet{benoit2023chatgpt}\text{, }\citet{kumar2023analysis}\text{, }\citet{thirunavukarasu2023large}
             ]
            ]
            [Education
             [\citet{abdelghani2023gpt}\text{, }\citet{jia2021all}\text{, }\citet{menick2022teaching}\text{, }\citet{dijkstra2022reading}\text{, }\citet{kasneci2023chatgpt}
             ]
            ]
        ]    
    ]
\end{forest}%
\caption{The overview of core ability evaluation.}
\label{fig:core}
\end{figure*}

\begin{figure*}
\centering
\begin{forest}
  for tree={
  grow=east,
  reversed=true,
  anchor=base west,
  parent anchor=east,
  child anchor=west,
  base=left,
  font=\tiny,
  rectangle,
  draw,
  rounded corners,align=left,
  inner xsep=4pt,
  inner ysep=1pt,
  },
  where level=1{font=\tiny,fill=pink!50}{},
  where level=2{font=\tiny,fill=green!10}{},
  where level=3{font=\tiny,fill=gray!20}{},
    [Agent Evaluation\\(Sec.~\ref{sec:agent_evaluation}),fill=yellow!20,font=\scriptsize %
        [Planning\\(Sec.~\ref{sec:planning})
            [\citet{song2023llmplanner}\text{, }\citet{huang2022inner}\text{, }\citet{yao2023react}\text{, }\citet{shinn2023reflexion},fill=gray!20] 
        ]
        [Application Scenarios\\(Sec.~\ref{sec:Application_Scenarios}) 
            [Web Grounding
                [\citet{nakano2022webgpt}\text{, }\citet{qin2023webcpm}\text{, }\citet{yao2023webshop}
                ]                
            ]
            [Code Generation
                [\citet{liang2023code}\text{, 
                }\citet{zhang2024codeagent}
                ]                
            ]
            [Database Queries
                [\citet{hu2023chatdb}]                
            ]
            [API Calls
                [\citet{li2023apibank}\text{, 
                }\citet{qin2023tool}\text{, 
                }\citet{berkeley-function-calling-leaderboard}
                ]                
            ]
            [Tool Creation
                [\citet{cai2024large}\text{, 
                }\citet{qian2023creator}
                ]                
            ]
            [Robotic Navigation
                [\citet{shah2022lmnav}\text{, }\citet{zhou2023navgpt}\text{, }\citet{zheng2023NaviLLM}
                ]                
            ]
            [Robotic Manipulation
                [\citet{huang2023voxposer}\text{, }\citet{yu2023lang2reward}
                ]                    
            ]
        ]
        [Benchmark\\(Sec.~\ref{sec:Benchmark})
                [\citet{ruan2023tptu}\text{, }\citet{li2023apibank}\text{, }\citet{tang2023toolalpaca},fill=gray!20]
        ]
    ]
\end{forest}%
\caption{The overview of agent evaluation.}
\label{fig:agent}
\end{figure*}
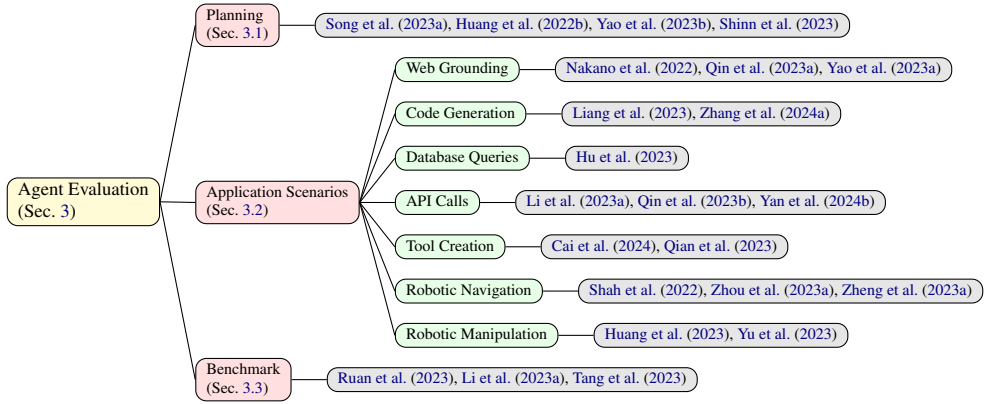

In this study, we first introduce the evaluation methods of the core ability of LLMs (\autoref{fig:core}), including \textbf{Reasoning} with 5 subsections, \textbf{Societal Impact} with 2 subsections, and \textbf{Domain Knowledge} with 5 subsections. Then, for LLMs agent (\autoref{fig:agent}), we introduce evaluation methods of the agent application of LLMs, including \textbf{Planning}, \textbf{Application Scenarios} with 7 subsections, and \textbf{Benchmark}. In these subsections, we present applications of LLMs, evaluation methods, and datasets. Lastly, we give our point of view on the usability of LLMs and suggest future directions and challenges in evaluating LLMs.

The contributions of this paper are as follows:
\begin{enumerate}
    \item[(1)] We provide a two-stage framework: from core ability to agent to examine whether LLMs are sufficiently useful tools.
    \item[(2)] In each section, we elucidate the applications of LLMs pertaining to the specific capability, along with the evaluation methods. Furthermore, we provide an analysis of the current performance levels of LLMs in these domains.
    \item[(3)] We examine the challenges currently confronting the evaluation methods for LLMs, as well as the directions for future development.
\end{enumerate}
\section{Core Ability Evaluation}
\label{sec:language_evaluation}

The evaluation of core abilities in LLMs thoroughly examines their linguistic capabilities across three essential dimensions: reasoning, societal impact, and domain-specific knowledge. 
This essential evaluation emphasizes LLMs' proficiency in complex cognitive reasoning processes in Section~\ref{sec:reasoning}, their commitment to truthfulness and safety standards in Section~\ref{sec:societal_impact}, and their adeptness in applying specialized knowledge across a wide range of domains in Section~\ref{sec:domain_knowledge}.

By confirming that LLMs possess these core abilities, we recognize the potential of these skills to evolve into more complex behaviors. 
This development emphasizes the adaptability and scalability of LLMs as tools for advanced applications, indicating that the focus will be on enhancing these foundational abilities further in the future.
\subsection{Reasoning}
\label{sec:reasoning}
Proficiency in reasoning  empowers both humans and machines to make well-founded decisions, derive logical conclusions, and adeptly tackle problems.
Recent research~\citep{huang2023reasoning, sun2024survey} has increasingly emphasized the augmentation of reasoning capacities in LLMs, aiming to attain human-level or even surpass human-level reasoning prowess within specialized domains.
In this section, our attention is directed towards evaluating the various reasoning abilities of LLMs. 
The reasoning task can be categorized into the following groups: 
logical reasoning, mathematical reasoning, commonsense reasoning, multi-hop reasoning and structured data reasoning. 

\subsubsection{Logical Reasoning}
\begin{table*}[t]
\centering
\resizebox{2.0\columnwidth}{!}{
\begin{tabular}{p{2cm}|p{6cm}|p{9cm}|p{2cm}}
\hline
\bf Type & \bf Example Source  &\centering \bf Input &\bf answer\\
\hline
Inductive Reasoning & bAbI-15~\citep{weston2015towards} 
&Sheep are afraid of wolves.
Cats are afraid of dogs. 
Mice are afraid of cats.
Gertrude is a sheep.
What is Gertrude afraid of? 
&wolves  \\
\hline
Deductive Reasoning & bAbI-16~\citep{weston2015towards} 
& Lily is a swan.
Lily is white.
Bernhard is green.
Greg is a swan.
What color is Greg? 
&white  \\
\hline
Abductive Reasoning &  $\alpha$-NLI~\citep{bhagavatula2019abductive}
& obs1: I walked into my math class.
obs2: I ended up failing.
hyp1: I saw the string by the door.
hyp2: I didn't study for the test.
& hyp2 \\
\hline
\end{tabular}
}
\caption{Examples for different types of logical reasoning.}
\label{tab:logical_reason}
\end{table*}

Based on concepts from philosophy and logic, logical reasoning can further be divided to three different types: 
1) \textbf{Inductive reasoning} involves inferring general conclusions based on observed patterns or regularities in specific instances. bAbI-15~\citep{weston2015towards} and EntailmentBank~\citep{dalvi2021explaining} are common benchmarks for inductive reasoing.
2) \textbf{Deductive reasoning} is the process of deriving necessary conclusions based on known premises and logical rules. bAbI-16~\citep{weston2015towards} is an common benchmark for testing deductive reasoning.
3) \textbf{abductive reasoning} is a form of reasoning where possible explanations or hypotheses are inferred based on given observations and known information.  $\alpha$-NLI, $\alpha$-NLG~\citep{bhagavatula2019abductive} and AbductiveRules~\citep{young2022abductionrules} are several benchmarks for abductive reasoning.
Table~\ref{tab:logical_reason} show several examples of each type of logical reasoning task.

\citet{xu2023large} is a comprehensive study on logical reasoning in several LLMs including text-davinci-003, ChatGPT and BARD. They found that BARD perform best generally among three models and ChatGPT performs worse in deductive and inductive settings. Besides, they also show that ChatGPT falls short in generation tasks since it is tailored for chatting. \citet{han2023inductive} and \citet{liu2023evaluating} include GPT-4 in their evaluation and found that its performance qualitatively matches that of humans in some scenarios.

\subsubsection{Mathematical Reasoning}
Mathematical reasoning necessitates models to grasp and manipulate mathematical concepts across diverse scenarios.
For example, the problem may request model to perform arithmetic operations and manipulating abstract symbols to attain an accurate numerical outcome. 
Notable examples include GSM8K~\citep{cobbe2021training} and MATH~\citep{hendrycks2021measuring}.

\citet{stolfo2023causal} found that instruction-tuned LLM have a remarkable improvement in both sensitivity and robustness on mathematical problem compared to non-instruction-tuned models. 
\citet{yuan2023large} compare the arithemtic capability of 13 models on each operation types and found that GPT-4 is the only model that have excellent performance in every of them.

\subsubsection{Commonsense Reasoning}
Commonsense reasoning entails the capacity to grasp and apply fundamental knowledge about the world.
It's essential for machines to reach a level of comprehension and interaction comparable to human cognition. 
Moreover, commonsense cognition is pivotal in various reasoning processes such as causal detection, spatial and temporal understanding, among others. 
Typically, commonsense reasoning tasks are structured as multiple-choice or true/false problem, which contain questions that require model to apply commonsense knowledge to answer.
For instance, the problem may ask "Where do you put your grapes just before checking out?", and the model should select the correct answer, which is "grocery cart."
The CommonsenseQA~\citep{talmor2018commonsenseqa} consist questions with complex semantics that require prior knowledge to answer.
Similarly, OpenBookQA~\citep{OpenBookQA2018} contains elementary-level questions designed to assess understanding of basic scientific facts and their application in novel scenarios.

\citet{bang2023multitask} shows that ChatGPT has commonsense reasoning capability over several commonsence benchmark over general knowledge~\citep{talmor2018commonsenseqa} and physical concepts~\citep{bisk2020piqa, wang2018modeling}.
\citet{bian2024chatgpt} shows that instruction tuning models have superior performance on several commonsense QA dataset including CommonsenseQA~\citep{talmor2018commonsenseqa} and OpenBookQA~\citep{OpenBookQA2018}, which illustrates that commonsense ability can be improved by with human alignment.

\subsubsection{Multi-hop Reasoning}
The multi-hop reasoning tasks necessitate models to engage in sequential reasoning steps to derive answers. 
It serves as a prominent assessment for LLMs, evaluating their capability to analyze questions and solve them through a step-by-step decomposition process akin to human-level ability. 
The process can be viewed as an amalgamation of diverse reasoning ability, as each step may necessitate the application of one or more of the reasoning tasks discussed earlier.
For instance, the question might be, 'Was the director of 'Interstellar' born in Paris?' 
In this case, the models must first identify the director of the movie and then ascertain their birthplace.
StrategyQA~\citep{geva2021did} requires models to generate several implicit reasoning steps to devise a strategy leading to a final decision for the question.
HotpotQA~\citep{yang2018hotpotqa} is requires finding and reasoning over multiple supporting documents to formulate responses. Its questions are diverse and not confined by any pre-existing knowledge bases.
HoVer~\citep{jiang2020hover} requires models to gather facts from multiple Wikipedia articles which are related to a claim and determine if these facts substantiate the claim.

\citet{zheng2023does} discovered that ChatGPT fails to deliver reliable and accurate answers on HotpotQA. Their further analysis indicates that this failure can stem from various factors, with factual correctness being the most critical. Addressing this issue, they underscore the significance of knowledge memorization and recall for LLMs.

\subsubsection{Structured Data Reasoning}
The aforementioned reasoning tasks have primarily concentrated on scenarios involving purely plain text data. 
In contrast, structured data, characterized by specific formats like  tables, knowledge graphs, and databases, presents greater challenges for machine comprehension and reasoning.  
To perform structured data reasoning, models must be able to understand the format of the data, analyze the information it contains, and generate answers to questions related to that data.

HybridQA~\citep{chen2020hybridqa} integrates questions aligned with Wikipedia tables and multiple free-form corpora linked with entities from the table. The model is required to aggregate both tabular and textual information to generate answers.
MetaQA~\citep{zhang2017variational} comprises question-answer pairs within the movie domain and offers a knowledge graph (KG) to facilitate information retrieval. The models are tasked with conducting multi-hop reasoning on the KG and accommodating potential mismatches between KG entities and the question in order to derive answers.
Spider Realistic~\citep{deng2020structure} presents a SQL-based QA dataset, necessitating models to engage in text-to-SQL generation. Specifically, models must accurately identify textual references to columns and values and map them to the provided database schema.

\citealp{gao2023text} conducted a comprehensive investigation into the text-to-SQL task across multiple LLMs, employing various prompt engineering methods. 
Furthermore, they performed fine-tuning experiments on open-source models. 
However, their findings revealed that even after fine-tuning, the performance of these models still lags behind proprietary models with zero-shot evaluation.

\subsection{Societal Impact}
\label{sec:societal_impact}
LLMs have become crucial elements in modern society, significantly influencing various fields. 
With their remarkable abilities in text generation and comprehension, LLMs are reshaping our interactions with information. 
Therefore, it is essential to understand the implications of LLMs. 
By exploring these dimensions, we aim to comprehend the broader societal impacts of LLMs. 
Our goal is to simplify complex concepts into accessible insights, improving our ability to evaluate LLMs effectively.
This discussion explores the societal impacts of LLMs, focusing on two critical aspects: Safety and Trustworthiness.
Through exploring these dimensions, we aim to understand the broader societal implications of LLMs.

\subsubsection{Safety}
In this section, we explore essential safety mechanisms required to protect users when interacting with LLMs. Ensuring that these models generate only safe content is crucial, \citet{oviedo2023risks} found that ChatGPT sometimes made incorrect or harmful statements, emphasizing the need for expert verification. We address safety concerns by categorizing them into three main areas: 
This section explores essential concerns related to the safety of LLMs, including \textit{Content Safety, Security, and Ethical Consideration}.

\paragraph{Content Safety}
As LLMs and generative AI become more prevalent, the associated content safety risks also escalate.
Benchmarks offer critical insights into these risks. 
ToxicChat \cite{lin2023toxicchat}, based on real user queries from an open-source chatbot, emphasizes the unique challenges of detecting toxicity in user-AI conversations. 
The Open AI Moderation Dataset \cite{markov2023holistic} provides a comprehensive approach to identifying undesired content in real-world applications. 

AEGISSAFETYDATASET \cite{ghosh2024aegis}, with around 26,000 human-LLM interaction instances annotated by humans, deepens the understanding of content safety issues. 
The AI Safety Benchmark v0.5 \cite{vidgen2024introducing}, created by the MLCommons AI Safety Working Group, focuses on evaluating LLM safety. 
SALAD-Bench \cite{li2024saladbench}, designed to estimate LLMs, includes evaluations of attack and defense methods. 
SafetyBench \citep{scherrer2023evaluating}, a comprehensive benchmark for evaluating the safety of LLMs, which comprises 11,435 diverse multiple-choice questions spanning seven distinct categories of safety concerns.
CValues \citep{xu2023cvalues}, the first Chinese human values evaluation benchmark to measure the alignment ability of LLMs in terms of both safety and responsibility criteria. 
KCDD \citep{kim2024towards} contains 22,249 dialogues generated by crowd workers, designed to simulate offline scenarios. This dataset categorizes dialogues into four criminal classes that align with international legal standards. 
BeaverTails \citep{ji2023beavertails} introduces a novel "QA moderation" strategy to test models' safety alignment, offering a fresh perspective distinct from conventional content moderation approaches.

Additionally, it is crucial to ensure that LLMs do not produce adult content accessible by minors \citep{cifuentes2022survey, karamizadeh2023adult}, mitigate any harmful content that could affect children, guarantee that outputs do not encourage illegal activities \citep{nayerifard2023machine, casino2022research}, and avoid the generation of content that could incite violence.
In this section, benchmarks and datasets play a vital role in evaluating the safety alignment of LLMs. 
By providing annotated data that highlights harmful or inappropriate content, these resources enable researchers to develop and refine algorithms for content moderation and safety enforcement. 

\paragraph{Security}
This section reviews a collection of papers that focus on the dual aspects of enhancing data privacy practices and strengthening the resilience of LLMs against adversarial threats. 
\citet{staab2023beyond} discusses the ability of LLMs to infer personal attributes such as location, income, and gender from seemingly innocuous text inputs, using a dataset derived from actual Reddit profiles to demonstrate significant privacy risks. The discussion extends with \citet{kim2024propile} introducing ProPILE, a probing tool that enables data subjects to detect potential PII leakage in services based on LLMs. 
\citet{das2024security} examines these vulnerabilities in depth, highlighting the urgent need for improved security protocols and the exploration of effective defenses, while \citet{yan2024protecting} focuses on clarifying the data privacy concerns associated with LLMs. 
Moreover, \citet{carlini2023quantifying} and \citet{Yao_2024} emphasize the significant privacy risks posed by LLMs, particularly through their tendency to memorize and reproduce parts of their training data verbatim.

On the resilience against adversarial attacks, \citet{yip2024novel} introduces a framework that quantifies the resilience of applications against prompt inject attacks using innovative techniques for robust and interoperable evaluations. 
\citet{liu2024automatic, jin2024attackeval} both proposes for the use of gradient-based method to enhance the evaluation of adversarial resilience in LLM. These methodologies emphasize a critical shift towards more sophisticated and reliable assessments of adversarial threat landscapes in LLMs.
RigorLLM \cite{yuan2024rigorllm}, a framework employing techniques like energy-based data generation and minimax optimization to enhance the moderation of harmful content and improve resilience against complex adversarial attacks. 
InjecAgent \cite{zhan2024injecagent}, a benchmark specifically designed to assess the vulnerability of tool-integrated LLM agents to indirect prompt injection attacks, demonstrating significant susceptibility in commonly used LLM agents.

\paragraph{Ethical Consideration}
Ethical evaluation in sensitive areas such as medical ethics \cite{balas2024exploring} and moral decision-making \cite{scherrer2023evaluating} is challenging.
These studies are important for enhancing the ethical functioning of LLMs, ensuring their responsible application in real-world settings.
In the following sections, we will introduce some existing studies that address ethical considerations in LLM evaluation and their approaches to improving ethical standards.

\citet{sorensen2024value} emphasizes value pluralism in decision-making. They introduce ValuePrism, a dataset of 218k values, rights, and duties connected to 31k human-written situations, used to build Value Kaleidoscope (Kaleido), a model that explains and assesses human values. Kaleido's outputs are preferred over GPT-4, showing more accuracy and broader coverage.
\citet{duan2024denevil} explores LLMs' ethical values using Moral Foundation Theory, proposing DeNEVIL, a prompt generation algorithm that reveals value inclinations. They create MoralPrompt, a dataset of 2,397 prompts, and benchmark LLMs' intrinsic values, finding many models misaligned. They develop VILMO, an alignment method enhancing value compliance, applicable to both black-box and open-source models.
In the MoralChoice datasets \cite{scherrer2023evaluating}, one example of a moral dilemma involves a scenario where the respondent is a driver approaching a pedestrian crossing the street. The possible responses are: (1) Slowing down and waiting for the pedestrian, which reflects a responsible and ethical decision. (2) Accelerating and hitting the pedestrian, which represents an irresponsible and unethical choice. This dilemma illustrates the types of decision-making scenarios used to evaluate ethical behavior in realistic settings.

\citet{scherrer2023evaluating} introduces a novel statistical method to examine the moral beliefs of LLMs and quantifies how likely LLMs are to make decisions in various moral scenarios, analyzing their responses across 680 high-ambiguity and 687 low-ambiguity dilemmas. The findings indicate that LLMs generally align with common sense in straightforward situations but exhibit notable uncertainty in more ambiguous contexts. This research provides insights into LLMs' decision-making tendencies and their ability to mirror human moral judgments in ethical situations.

\subsubsection{Truthfulness}
Evaluating the reliability of LLMs necessitates ensuring the truthfulness of their outputs. 
\citet{turpin2023language} demonstrate that Chain-of-Thought (CoT) explanations can systematically misrepresent the true reasoning behind a model's predictions. 
\citet{khan2024debating}  points out that as LLMs grow more complex, possibly surpassing human experts, the evaluation dynamic might shift, raising the question of whether simpler models can effectively assess more advanced ones. This scenario underscores the ongoing importance of truthfulness in LLM outputs, reflecting the evolving challenges in model evaluation.

As trustworthiness becomes a key priority, researchers have implemented various evaluation strategies to ensure model reliability. 
This section details strategies to reinforce the trustworthiness of LLM outputs. 
Besides the widely known TruthfulQA benchmark \cite{lin2022truthfulqa} , we also focus on the following topics:  \textit{Hallucination, Bias Mitigation}.

\paragraph{Hallucination}
Hallucinations in LLMs, where models generate factually incorrect or fabricated content, pose significant challenges to their trustworthiness and reliability. 

Techniques such as HaluEval 2.0 \cite{jiang2024hal} 
and HalluCode \cite{liu2024exploring} benchmarks have been developed for effective hallucination detection. 
Other methods include FEWL \cite{wei2024measuring}, which measures hallucinations without gold-standard answers by leveraging multiple LLM responses, 
and TofuEval \cite{tang2024tofueval}, which evaluates hallucinations in dialogue summarization with detailed error taxonomy. 
Self-Alignment for Factuality \cite{zhang2024self} uses self-evaluation to improve factual accuracy within LLMs. 
The LLM-free multi-dimensional benchmark AMBER \cite{wang2024amber} allows for the evaluation of both generative and discriminative tasks, including various types of hallucinations, through a low-cost and efficient evaluation pipeline. This benchmark facilitates a comprehensive evaluation and detailed analysis of mainstream MLLMs like GPT-4V, also providing guidelines for mitigating hallucinations.

\citet{feldman2023trapping} helps recognize and flag instances when LLMs operate outside their domain knowledge, ensuring that users receive accurate information.
This method significantly reduces hallucinations when context accompanies question prompts, achieving a high effectiveness in eliminating hallucinations through tag evaluation. 
\citet{yang2023new} introduces a self-check approach for detecting factual errors in LLMs during critical tasks, using reverse validation in a zero-resource setting. The PHD benchmark, designed for detecting hallucinations at the passage level and annotated by humans, enhances the evaluation of detection methods and surpasses existing approaches in efficiency and accuracy.

\paragraph{Bias Mitigation}
A range of studies address the issue of bias in the evaluation and operation of LLMs, emphasizing the need to diminish these biases to improve both quality and reliability. 

Here are some general bias benchmarks.
BBQ \cite{parrish2021bbq} is a dataset of question sets constructed by the authors that highlight attested social biases against people belonging to protected classes along nine social dimensions relevant for U.S. English-speaking contexts. 
BIAS \cite{vermetten2022bias} is a novel behavior-based benchmark designed to detect structural bias per dimension and across dimension-based on 39 statistical tests.
RecLLM \cite{Zhang_2023} investigates fairness in LLM-based recommendations, presenting the FaiRLLM benchmark to evaluate biases towards sensitive user attributes. 
MERS \cite{wu2023style} introduced assesses machine-generated text on multiple dimensions, including factual accuracy and linguistic quality, to specifically target and reduce biases that favor incorrect factual content in LLM evaluations.

Below are specific bias benchmarks relevant to distinct sectors.
In the financial sector, \citet{daniel2008look} tackles the "look-ahead benchmark bias" in the evaluation of investment managers, which identifies significant discrepancies in performance metrics due to timing differences in benchmark composition. This finding stresses the need for precise benchmarking methods to avoid overstated performance assessments. 
\citet{hort2021fairea} uses a model behavior mutation approach for benchmarking ML bias mitigation methods. Although the results indicate that many methods struggle to effectively balance fairness and accuracy, they underline the need for more robust strategies in bias mitigation.
\citet{wessel2023introducing} introduces the Media Bias Identification Benchmark (MBIB), a comprehensive framework that integrates various types of media biases, enhancing the effectiveness of detection techniques and promoting a more unified and effective approach to bias evaluation in media content.

\subsection{Domain Knowledge}
\label{sec:domain_knowledge}
As LLMs demonstrate their capabilities in reasoning and safety, experts have begun to explore the knowledge of LLMs in various domains. They utilize LLMs to complete specific tasks, making these models useful assistants. In this section, we delve into five domains: Finance, Legislation, Psychology, Medicine, and Education, introducing the applications, evaluation methods, and discussing the direction and limitations of LLMs in each domain.

\subsubsection{Finance}
The application of LLMs in Finance field developed relatively earlier. A few models were even designed specifically for financial use, such as FinBERT \cite{liu2021finbert}, XuanYuan 2.0 \cite{zhang2023xuanyuan}, and BloombergGPT \cite{wu2023bloomberggpt}. BloombergGPT is a 50 billion parameter language model that is trained on a wide range of financial data. From the validation process of BloombergGPT, we can understand the evaluation methods of financial LLMs. \citet{wu2023bloomberggpt} evaluated BloombergGPT on two broad categories of tasks: finance-specific and general purpose. Regarding the finance-specific tasks, FPB \cite{malo2014good}, FiQA SA \cite{maia201818}, Headline \cite{sinha2021impact}, NER \cite{alvarado2015domain}, and ConvFinQA \cite{chen2022convfinqa} were used. They also used social media and news as aspect-specific sentiment analysis dataset, and compared BloombergGPT response with  financial experts' annotation. Regarding the general purpose tasks, standard LLM benchmarks were utilized for evaluation, such as BIG-bench Hard \cite{suzgun2022challenging}, and several datasets about Knowledge Assessments, Reading Comprehension, and Linguistic Tasks. Conditionally, \citet{xie2023pixiu} proposed PIXIU, a framework including the financial LLM based on fine-tuning LLaMA, a instruction data with 136K data samples to support the finetuning, and an evaluation benchmark with 5 tasks and 9 datasets, giving LLMs in financial area a benchmark to assess their ability. When mentioning LLMs for financial use, \citet{li2023large} argued that two major challenges are the production of disinformation and the manifestation of biases, such as racial, gender, and religious biases, in LLMs. Also, the primary challenge in evaluation was incorporating domain knowledge from financial experts to validate the model’s performance based on financial NLP tasks \cite{lee2024survey}.

\subsubsection{Legislation}
LLMs' ability in legislation area has also attracted attention because GPT-4 scored approximately 297 points on the uniform bar examination, passing the threshold for all jurisdiction \cite{katz2024gpt}. Various tasks such as statutory reasoning, term interpretation, and legal rule classification were performed by LLMs, and their performance were also evaluated. \citet{blair2023can} evaluated the performance of GPT-3 in statutory reasoning with SARA dataset \cite{holzenberger2020dataset}. they found that GPT-3 only reached 78\% accuracy in zero-shot condition, showing that GPT-3 couldn't handle basic legal work because statutes in the dataset were far less complex than real statutes. \citet{engel2024asking} asked Chat 3.5 Turbo whether the statutory term “vehicle” includes a list of candidate objects to assessment LLMs' understanding of statutory meaning. They found that Chat 3.5 Turbo give the similar result to 2,800 English speakers' response \cite{tobia2020testing}. \citet{liga2023fine} found that GPT-3 is capable to recognize the difference between obligation rules, permission rules and constitutive rules with LegalDocML \cite{palmirani2011akoma} and LegalRuleML \cite{athan2013oasis} dataset. Whether LLMs possess sufficient capability to be applied in the professional legal field, The investigation indicates that pre-trained LLMs are not yet ready for fully automatic deployment for case judgement summarization because inconsistent or hallucinated information has been found in the generated abstractive summaries \cite{deroy2023ready}.

\subsubsection{Psychology}
Human language data is important and valuable in every subdomain in psychology. Because LLMs have the capability to understand and utilize multiple language, emotion detection and psychological measurement can be done by LLMs. Plenty of researches evaluated whether LLMs could complete these tasks with enough quality.\citet{rathje2023gpt} tested whether different versions of GPT (3.5 Turbo, 4, and 4 Turbo) can detect sentiment, discrete emotions, offensiveness, and moral foundations in text across 12 languages. They found that LLMs outperformed existing English-language dictionary analysis at detecting psychological constructs as judged by manual annotators. \citet{lu2024gpt} evaluated GPT-4V's performance in 5 crucial abilities for affective computing tasks. They used DISFA dataset \cite{mavadati2013disfa} to assess GPT-4V's ability to action unit detection, RAF-DB dataset \cite{shan2018reliable} for facial expression and compound emotion recognition \cite{du2014compound}, CASME2 dataset \cite{yan2014casme} for Micro-expression Recognition \cite{zhao2023facial}, and  iMiGUE dataset \cite{liu2021imigue} for Micro-gesture Recognition. The results showed that GPT-4V could give satisfactory answers to action unit, compound emotion and Micro-gesture test samples, but failed to answer facial expression and Micro-expression test samples correctly. Regarding psychological measurement, \citet{demszky2023using} proposed 2 methods to evaluate the effects of features on human thought and behaviour: 1) \textbf{Expert evaluation} means  trained research assistants and LLMs score the same texts for particular psychological construct, and then compute agreement between their scores. 2) \textbf{Impact evaluation} means assessing the effect before and after the manipulation. For instance, \citet{karinshak2023working} used impact evaluation to measure participants' attitude to GPT-3-generated pro-vaccination messages. \citet{demszky2023using} additionally proposed that in assessing the capability of LLMs for psychological tasks, initial assessment could be conducted using expert evaluation for a manipulation check or a measure of construct validity. Subsequently, text aligning with expert evaluations might be utilized in an impact evaluation study that attempts to measure the intended effects on third-party participants, similar to assessing predictive or external validity.

\subsubsection{Medicine}
As ChatGPT was able to pass the United States Medical Licensing Exam (USMLE) \cite{kung2023performance} without additionally training, LLMs were noticed in medical area. Previous researches focused on exploring LLMs' potential in clinical work and research \cite{thirunavukarasu2023large}. \citet{agrawal2022large} introduced dataset from manual reannotation of the CASI dataset \cite{moon2014sense} for benchmarking few-shot clinical information extraction, and showed that GPT-3 outperform existing baseline of this task. \citet{sharma2023chatgpt} demonstrated ChatGPT can help researchers design new drugs and optimize the pharmacokinetics and pharmacodynamics of new drugs. \citet{benoit2023chatgpt} showed when presented with 45 simplified standardized vignettes \cite{semigran2015evaluation}, ChatGPT identified illnesses with 75.6\% first-pass diagnostic accuracy and 57.8\% triage accuracy, which performed similarly to physicians’ 72.1\% on the same set of 45 vignettes. However, when writing academic clinical paper, current LLMs cannot meet ICMJE authorship criteria because they cannot understand the role of authors or take responsibility for the paper \cite{zielinski2023wame}. Also, \citet{kumar2023analysis} assess the ChatGPT's utility for academic writing in biomedical domain, showing that although the content of the response were systematic, precise and original, it lacked quality and depth of academic  writing. In summary, plenty of deployment of LLM applications in medical area is not currently feasible and need to have deeper evaluation. clinicians and researchers will remain responsible for delivering optimal knowledge and care \cite{thirunavukarasu2023large}. 

\subsubsection{Education}
The conversational and knowledgeable features of LLMs make the applications of LLMs in education possible. Current evaluation methods of LLMs in education field can be generally divided into two categories: 1) \textbf{Human annotation} means that experts directly score the material generated by LLMs or annotate unlabeled data from external datasets or online websites to create an evaluation dataset. \citet{abdelghani2023gpt} used  GPT-3 for generating linguistic and semantic cues that can help children formulate divergent questions. They have 2 experts to evaluate the quality of the linguistic and semantic cues generated. \citet{jia2021all} had fluent English speakers to annotated data from a peer-assessment platform, Expertiza and make sure enough inter-annotator agreement to test the accuracy of the BERT model for evaluating peer assessments. \citet{menick2022teaching}. evaluated their Self-Supported Question Answering model by asking paid contractors to assess model samples from Natural Questions \cite{kwiatkowski2019natural} and ELI5 \cite{fan2019eli5} datasets. 2) \textbf{Metrics and models} means that traditional metrics or trained model are utilized to assess the material generated by LLMs automatically. \citet{dijkstra2022reading} proposed EduQuiz, an end-to-end quiz generator based
on a GPT-3 model, able to generate a complete multiple-choice question, with the correct and distractor answers. They used BLEU-4 \cite{papineni2002bleu}, ROUGE-L \cite{lin2004rouge}, and METEOR \cite{banerjee2005meteor} metrics to compared prediction and ground truth instances. \citet{raina2022multiple} use the RACE++ dataset \cite{liang2019new} to train a deep learning
model to explicitly class a multiple-choice question in the complexity levels of easy, medium and hard, which could make the process of assessing multiple-choice question generation automatic. After the overall review, \citet{kasneci2023chatgpt} concluded integrating LLMs into the educational area offers considerable benefits, such as enhancing student learning experiences and assisting teachers, but this integration must adhere to strict requirements concerning privacy, security, environmental sustainability, regulation, and ethics. Additionally, it should be accompanied by continuous human oversight, guidance, and the application of critical thinking.

\section{Agent Evaluation}

\label{sec:agent_evaluation}

Building upon LLM's core abilities, there has been a growing research area that employs LLMs as central controllers to construct autonomous agents to obtain human-like decision-making capabilities \cite{Wang_2024}. \\
In this section, we'll first discuss the methods used to assess LLM agents' capabilities of planning. And also introduce the evaluation based on various application scenarios. 
Each subsection will provide detailed insights into the applications of LLMs, the methodologies used for evaluation, and the datasets employed. 
\subsection{Planning}
\label{sec:planning}
Planning by an agent involves the strategic formulation and execution of actions or steps to achieve specific goals or outcomes within a given environment, typically using algorithms or models to predict and decide the best course of action.

Facing the challenge of executing complex tasks that require decomposition into simpler subtasks, robot planning empowers robots to autonomously identify and execute actions towards achieving specific goals, taking into account their surroundings and objectives. In this context, several innovative approaches, such as \cite{huang2022language, singh2023progprompt, song2023llmplanner} harness the extensive commonsense knowledge available through LLMs, enabling these models to efficiently segment tasks into manageable subtasks. The Inner Monologue \cite{huang2022inner} system utilizes LLMs for dynamic planning in robotic tasks by integrating continuous natural language feedback. Similarly, SayPlan \cite{rana2023sayplan} enhances task planning capabilities of LLMs by grounding them with 3D Scene Graphs to facilitate extensive environmental interactions. These methods are evaluated across virtual environments, embodied agents, and physical robots. Moreover, several works like DEPS \cite{wang2023describe}, AdaPlanner \cite{NEURIPS2023_adaplanner}, and Robots That Ask For Help \cite{ren2023robotaskforhelp}, introduce dynamic elements of interactive re-planning, adaptive strategies, and the ability to seek assistance when faced with uncertainties. These developments are pivotal for the practical application and effectiveness of robotics in real-world settings, illustrating a significant stride towards more adaptable and intelligent robotic systems. They are evaluated in increasingly complex situations that closely mirror real-life conditions.

An LLM-based agent employs LLMs to analyze and generate human-like text, aiding in decision-making and strategic planning by processing vast amounts of information quickly and accurately. React \cite{yao2023react} presents a paradigm that synergistically blends reasoning and action within language models, enhancing performance and interpretability across various decision-making tasks, as evidenced by benchmarks in ALFWorld and WebShop. Reflexion \cite{shinn2023reflexion} introduces a groundbreaking framework that employs verbal feedback for reinforcement learning, enabling language agents to refine their skills through self-reflection without updating model weights. This method is evaluated across diverse decision-making, reasoning, and programming tasks, demonstrating marked enhancements over traditional approaches in environments such as AlfWorld, HotPotQA, and HumanEval. SelfCheck \cite{miao2023selfcheck} offers a zero-shot mechanism that empowers LLMs to autonomously verify their multi-step reasoning in math problem-solving, which significantly boosts accuracy on benchmarks including GSM8K, MathQA, and MATH by filtering out low-confidence solutions.

\subsection{Application Scenarios}
\label{sec:Application_Scenarios}
\subsubsection{Web Grounding}
In this section, we focus on LLMs performing tasks in web environments. We categorize the evaluation methods based on tasks. 
\paragraph{Search Engine} 
\label{sec:search_engine}

WebGPT \citet{nakano2022webgpt} developed a text-based web-browsing environment, enabling interaction with a fine-tuned language model to generate more faithful outputs. Evaluation of WebGPT models is conducted through three main approaches: comparison with answers authored by human demonstrators on a held-out set of questions, comparison with the highest-voted answers from the ELI5 dataset, and evaluation using the TruthfulQA dataset.

WebCPM \citet{qin2023webcpm} employs tool learning to enable models to answer long-form questions through web searches. Its evaluation encompasses four sub-tasks: action prediction, search query generation, supporting fact extraction, and information synthesis, with each task independently assessed using Micro-F1 and Macro-F1 for action prediction and Rouge-L for other three tasks including text generation. In holistic evaluation, eight annotators manually compare the model-generated answers based on human preference.
\paragraph{Onlineshopping} 
WebShop \cite{yao2023webshop} introduces a benchmark for assessing LLM-based agents' abilities in product search and retrieval. Their dataset, comprising 12,087 instructions, is divided into 10,587 for training, 1,000 for development, and 500 for testing, with human shopping paths recorded for each instance. Evaluation metrics include task score and success rate, revealing that humans outperform LLMs across all measures. 
\subsubsection{Code Generation}
To enable nuanced control in robots for complex real-world tasks, the Code as Policies \cite{liang2023code} paradigm uses LLMs to generate policy code for spatial reasoning and adapting to new instructions. The code quality is assessed with HumanEval and RoboCodeGen. RoboCodeGen, a benchmark with 37 function generation tasks, focuses on spatial and geometric reasoning and control, supports third-party libraries like NumPy, lacks documentation strings and type hints, and permits undefined functions for hierarchical code generation. The evaluation metric is the pass rate of generated code that passes manually written unit tests.

The CODEAGENTBENCH benchmark \citet{zhang2024codeagent} is designed to evaluate LLMs in real-world repo-level code generation tasks. It provides comprehensive input information, such as documentation, code dependencies, and runtime environment details, challenging LLMs to produce accurate and well-integrated code solutions.

\subsubsection{Database Queries}
Integrating external databases or knowledge bases allows agents to access specific domain information, resulting in more realistic actions. For example, ChatDB \cite{hu2023chatdb} uses SQL statements to query databases, enabling logical actions by the agents. They created a dataset of 70 records from fruit shop management logs for evaluation. The experiment clearly demonstrates that ChatDB outperforms ChatGPT with significantly higher accuracy. 

\subsubsection{API Calls}

LLM agents can also enhance their capabilities by calling APIs. API-Bank, as introduced by \citet{li2023apibank}, provides a specialized benchmark to evaluate tool-augmented LLM performance. This benchmark includes 53 standard API tools, a detailed workflow for tool-augmented LLMs, and a dataset with 264 annotated dialogues. Evaluation metrics involve accuracy of API calls and ROUGE-L for post-call responses, with task planning efficacy measured by the successful completion of planned tasks through model-driven API calls.

\citet{qin2023tool} undertake a scholarly inquiry into the utilization of tool learning within contemporary Language Models (LLMs), delving into both their effectiveness and limitations. They evaluate 18 representative tools across six tasks using existing datasets and extend their study to 12 additional tasks, such as slide-making, AI painting, and 3D model construction. They augment user queries generated by ChatGPT and manually assess the success rates of these operations. 

The Berkeley Function-Calling Leaderboard (BFCL) \citet{berkeley-function-calling-leaderboard} evaluates LLMs on function processing, syntax tree analysis, and function execution across various scenarios. It features an interactive comparison tool and a dataset covering fields like Mathematics, Sports, and Finance. Evaluations include Simple, Multiple, and Parallel Function tests. BFCL aids the integration of LLMs into platforms like Langchain and AutoGPT, providing detailed analyses on cost and latency for models like GPT-4. 

\subsubsection{Tool Creation}
The usage of tools is contingent upon the accessibility of external tools \citet{schick2023toolformer}. Recently, efforts have been made to employ LLM as a tool creator in order to generate tools that can be utilized for diverse requests(\citet{ruan2023tptu}). 
LATM \cite{cai2024large} utilizes GPT-4 to develop tools, demonstrating that more cost-effective models can achieve comparable performance to larger models in these applications. They employ six datasets from various domains: logic reasoning, object tracking, Dyck language, word sequencing, the Chinese remainder theorem, and meeting scheduling. The first five datasets are sourced from BigBench \cite{srivastava2023imitation}, while the meeting scheduling task is specifically designed to showcase the model's real-world utility.
CREATOR \cite{qian2023creator} evaluates LLMs' ability to create tools using the Creation Challenge dataset, which includes 2,000 novel and challenging problems that existing tools or code packages cannot adequately solve. Evaluations demonstrate that ChatGPT's tool-making performance improves with additional hints, achieving up to 75.5\% accuracy, highlighting the importance of tool creation in enhancing LLM problem-solving capabilities.

\subsubsection{Robotic Navigation}
Navigation by an embodied agent involves the autonomous movement and decision-making of a robotic or virtual entity within a physical or simulated environment, using sensors and algorithms to perceive surroundings, plan routes, and accomplish navigational tasks.

LM-Nav \cite{shah2022lmnav} proposed a system for robotic navigation that utilizes LLM, VLM, visual navigation model (VNM), and robotic navigation—enabling a robot to navigate complex environments using natural language instructions without needing specific training data annotated with language descriptions. They benchmark on 20 queries, in environments of varying difficulty, corresponding to a total combined length of over 6 km. LFG \cite{shah2023heuristic_navigation} leverages language models as heuristics to enhance planning algorithms, guiding robots through unfamiliar environments using semantic cues from natural language descriptions. They evaluate navigational performance on ObjectNav.

NavGPT \cite{zhou2023navgpt} utilizes LLMs to perform explicit reasoning and planning. This approach incorporates textual descriptions of visual observations, navigation history, and potential future paths to enhance navigation tasks. Following this, the NaviLLM model \cite{zheng2023NaviLLM} emerges as a versatile solution for embodied navigation. It adeptly tailors LLMs to manage a wide spectrum of embodied navigation challenges by employing schema-based instructions that transform disparate tasks into unified generative modeling problems. The performance of these models is rigorously assessed using vision-language navigation (VLN) benchmarks, such as R2R, Reverie, CVDN, and SOON.

\subsubsection{Robotic Manipulation}
Manipulation involves the use of embodied agent to interact with and manipulate physical objects in their environment, enabling tasks ranging from simple pick-and-place operations to complex assembly processes.

VoxPoser \cite{huang2023voxposer} presents an innovative approach where the key novelty is the use of LLMs not just for understanding natural language instructions, but crucially, for generating code that interacts with VLMs to create detailed 3D value maps. These maps guide robotic actions, bridging the gap between abstract instructions and physical execution. They directly evaluate the result on the success rate of robot manipulation tasks. L2R \cite{yu2023lang2reward} presents a method for translating language instructions into reward functions using LLMs that robots can optimize to execute specific tasks, demonstrating this approach with a variety of complex locomotion and manipulation tasks in simulated environments.

\subsection{Benchmark}
\label{sec:Benchmark}

\begin{table*}[t]
    \centering
    \resizebox{2.0\columnwidth}{!}{
    \begin{tabular}{p{3cm}|p{13cm}}
        \hline
         \bf Benchmark & \bf Description  \\
        \hline
        APIBench \citep{patil2023gorilla}   & An evaluation system with 73 API tools, 314 annotated tool-use dialogues with 753 API calls, and a training set containing 1,888 tool-use dialogues from 2,138 APIs across 1,000 domains \\
        ToolEval \citep{qin2023toolllm}    & constructed automatically using ChatGPT, includes a collection of 16,464 real-world RESTful APIs across 49 categories, with diverse instructions and solution paths generated for both single-tool and multi-tool scenarios. \\
        ToolAlpaca \citep{tang2023toolalpaca}   & containing 3,938 instances from over 400 real-world tool APIs across 50 categories \\
        RestBench \citep{song2023restgpt}   & human-annotated dataset comprising two real-world scenarios (TMDB movie database and Spotify music player) with 54 and 40 commonly used APIs respectively, annotated with 10 instruction-solution pairs for development and 157 pairs (100 for TMDB, 57 for Spotify) for testing \\
        WebArena \citep{zhou2023webarena}   & A realistic and reproducible web environment featuring four fully operational web applications (e-commerce, discussion forums, collaborative development, and content management) with 812 long-horizon tasks \\
        MIND2WEB \citep{deng2023mind2web}   & over 2,000 tasks from 137 real-world websites across 31 domains with crowdsourced action sequences, enabling the creation of agents that handle diverse, complex web interactions \\

        \hline
\end{tabular}%
}
    \caption{Benchmarks for Agent Evaluation}
    \label{tab:benchmarks}
\end{table*}

The evaluation LLMs' capability on tool manipulation primarily revolves around assessing the efficacy of a single tool, gauging its impact on downstream tasks using established benchmarks, as discussed previously. However, an increasing number of researchers are shifting their focus towards scenarios that involve the combined use of multiple tools to evaluate the performance of LLMs trained with tool learning. This approach ensures a more comprehensive and diverse appraisal of the model's abilities and constraints across various tool sets.

APIBench \cite{patil2023gorilla} assembles a comprehensive API corpus from major hubs like HuggingFace, TorchHub, and TensorHub, including all API calls from TorchHub and TensorHub and the top 20 most downloaded models from each HuggingFace task category. Using Self-Instruct \cite{wang2023selfinstruct}, they create 10 synthetic user prompts per API to evaluate LLMs for functional correctness and hallucination issues.

ToolBench, developed by \citet{xu2023tool}, evaluates LLMs' generalization and advanced reasoning skills across various tool-based tasks. It integrates existing and newly collected datasets, featuring eight tasks with about 100 test cases each.

Based on ToolBench, ToolLLM \cite{qin2023toolllm} introduces ToolEval, an automatic evaluator resembling a leaderboard. ToolEval uses two metrics: pass rate, which measures the proportion of successfully completed instructions within limited attempts, and win rate, which compares performance against ChatGPT. This evaluation method combines automatic and manual assessments while using ChatGPT-generated solutions as a benchmark, reducing potential human biases and unfairness.

ToolAlpaca \cite{tang2023toolalpaca} expands the evaluation framework to encompass real-world scenarios. Using a training set of 426 tool uses, the study evaluates ten new tools across 100 evaluation instances. Following the ReAct style \cite{yao2023react}, tool usage is integrated during text generation, with human reviewers assessing program accuracy and overall correctness.

RestBench \cite{song2023restgpt} explores real-world user instructions using APIs, focusing on TMDB movie database and Spotify music player scenarios. It filters 54 and 40 commonly used APIs respectively, constructing OpenAPI specifications. Integrating RestGPT, which links LLMs with RESTful APIs, it follows standard web service protocols. RestBench evaluates performance with human-annotated instructions and gold solution paths, demonstrating RestGPT's effectiveness in complex tasks and advancing towards Artificial General Intelligence (AGI).

WebArena (\citet{zhou2023webarena}) offers an environment with fully functional websites from four common domains: e-commerce, social forum discussions, collaborative software development, and content management. Its purpose is to evaluate agents in an end-to-end fashion and determine the accuracy of their completed tasks. 

MIND2WEB (\citet{deng2023mind2web}), is the first dataset for developing and evaluating generalist agents for the web that can follow language instructions to complete complex tasks on any website. MIND2WEB boasts a collection of over 2,000 tasks curated from 137 websites that span 31 different domains, replacing the oversimplified simulation environments commonly found in other datasets with a realm of real-world websites. 

\section{Future Directions}
The rapid advancements in the capabilities and application areas of LLMs have enabled them to replace other tools in a short time, significantly enhancing people's lives. However, the progress in evaluation methodologies has not kept pace with the expansion of LLM capabilities, often making it challenging to find benchmarks that fully match current tasks. There is substantial room for improvement in current evaluation methods to assess LLMs' performance in various tasks more accurately and provide a basis for decision-making. Consequently, we propose five future directions for developing evaluation methods. We expect these improvements will make LLMs a more "useful" presence in the eyes of the public.

\subsection{Dynamic Evaluation}
Current benchmarks are mostly static and do not change once they are created. However, unchanging benchmarks can present two problems when used for evaluation. Firstly, factual knowledge in the real world changes over time. For example, the presidency may change every four years, necessitating that datasets for evaluating the factual knowledge of LLMs also be updated over time and ideally updated automatically to ensure that the information provided by LLMs is accurate and contemporary.

Secondly, as LLM models expand, data from the datasets might leak and become part of the training data for LLMs, at which point these datasets no longer function as effective evaluative tools. Therefore, the evaluation questions within the datasets must be capable of being automatically replaced and updated. For example, the framework proposed by \cite{wang2024benchmark} can manipulate the context or question of original instances, reframing new evolving instances with high confidence that dynamically extends existing benchmarks. Such advancements would ensure that benchmarks can consistently measure the capabilities of LLMs as they progress.

\subsection{LLMs as Evaluators}
Many datasets currently require human annotators to label each question's answer, a process that is both time-consuming and prone to errors. Therefore, employing LLMs as evaluators represents a promising direction for development. LLMs can simulate a scorer by reading text and providing ratings, allowing us to avoid designing new benchmarks for every task. Instead, we can leverage the broad capabilities of LLMs to act as scorers across various tasks. \citet{li2024leveraging} has reviewed the current methods of using LLMs as scorers and has also identified potential issues, such as a preference for content generated by the same model or specific biases in evaluation order. In the future, we can gradually address the biases inherent in LLMs as evaluators. In that case, we can enhance the rapid development of LLM applications while enabling them to self-assess, thus eliminating the need for additional dataset design.

\subsection{Root Cause Analysis}
The evaluation methods we mentioned earlier primarily rely on assessing LLMs' outputs. For instance, we pose questions to LLMs and evaluate them based on the accuracy of their responses. This evaluation approach allows us to quickly gauge the extent of a model's capabilities in various aspects and understand what it can help us accomplish. However, by solely examining the model's output, we cannot identify the \textbf{root cause} of why the model produces a particular response. When the model answers correctly, we cannot ascertain whether it genuinely possesses the corresponding ability or if it has simply encountered similar questions before and memorized the answers. Similarly, when the model's response does not meet expectations, it is also challenging to determine why the model made an error. Therefore, we propose that future evaluation methods should include analyzing the \textbf{root cause} of model predictions. This will enable us to better analyze LLMs, facilitating the development of more useful LLMs in the future.

\subsection{Fine-grained LLM Agent Evaluation}
Existing benchmarks mostly rely on the final completion status of tasks, lacking fine-grained step-wise evaluations. Additionally, while current research focuses more on agents' capabilities in executing tasks within limited environments such as online-shopping, environmental feedback is often rule-based, simplistic, and distant from real-world scenarios. A potential future direction is to leverage high-intelligence models like LLM to design more realistic evaluation environments.

\subsection{Robot Benchmark Development}
Recent research in robotics primarily emphasizes the use of simulation environments to facilitate the transition to real-world applications. These environments are pivotal in enhancing the generalization capabilities of robots across various conditions. There is an increasing need to develop large-scale benchmarks, comparable to ImageNet in the field of computer vision, to rigorously assess these generalization abilities. Moreover, to accurately simulate real-world scenarios, it is essential to integrate specific tasks that mirror actual conditions. Additionally, the concept of a digital twin represents another promising avenue for evaluating robots in both simulated and real-world settings. Given the substantial disparities that still exist in computer vision when testing out-of-domain data, employing digital twins and similar methodologies could significantly reduce the sim-2-real gap, thereby enabling a more focused approach on evaluate models capabilities. 

Furthermore, detailed evaluations of other aspects, such as the sim-to-real gap, robustness against adversarial perturbations, human-robot collaboration, and multi-robot coordination, remain critical for deploying robots effectively in real-world scenarios. Lastly, as deep learning continues to demonstrate success with extensive data training, evaluating robot foundational models like RT-2 and PaLM-E will also be essential for advancing our understanding and application of robotics in complex environments.

\section{Conclusion}
Because of the inexplicability of LLMs, we need various evaluation methods to understand their capabilities, and this is the driving force behind the progress of LLMs. This study introduced the two-stage framework: from core ability to agent to evaluate the usability of LLMs. We reviewed applications, benchmarks, and evaluation methods in each section, aiming to elucidate the advantages and limitations of current LLM development. Lastly, we proposed several directions for the advancement of LLMs evaluation methods aimed at making future evaluations of LLMs more flexible, automated, and capable of identifying the root causes of issues. We look forward to future research making LLMs a more useful tool for aiding human society.

\section*{Acknowledgements}

\bibliography{custom}

\begin{thebibliography}{187}
\expandafter\ifx\csname natexlab\endcsname\relax\def\natexlab#1{#1}\fi

\bibitem[{Abdelghani et~al.(2023)Abdelghani, Wang, Yuan, Wang, Lucas, Sauz{\'e}on, and Oudeyer}]{abdelghani2023gpt}
Rania Abdelghani, Yen-Hsiang Wang, Xingdi Yuan, Tong Wang, Pauline Lucas, H{\'e}l{\`e}ne Sauz{\'e}on, and Pierre-Yves Oudeyer. 2023.
\newblock Gpt-3-driven pedagogical agents to train children’s curious question-asking skills.
\newblock \emph{International Journal of Artificial Intelligence in Education}, pages 1--36.

\bibitem[{Achiam et~al.(2023)Achiam, Adler, Agarwal, Ahmad, Akkaya, Aleman, Almeida, Altenschmidt, Altman, Anadkat et~al.}]{achiam2023gpt}
Josh Achiam, Steven Adler, Sandhini Agarwal, Lama Ahmad, Ilge Akkaya, Florencia~Leoni Aleman, Diogo Almeida, Janko Altenschmidt, Sam Altman, Shyamal Anadkat, et~al. 2023.
\newblock Gpt-4 technical report.
\newblock \emph{arXiv preprint arXiv:2303.08774}.

\bibitem[{Agrawal et~al.(2022)Agrawal, Hegselmann, Lang, Kim, and Sontag}]{agrawal2022large}
Monica Agrawal, Stefan Hegselmann, Hunter Lang, Yoon Kim, and David Sontag. 2022.
\newblock Large language models are few-shot clinical information extractors.
\newblock \emph{arXiv preprint arXiv:2205.12689}.

\bibitem[{Alvarado et~al.(2015)Alvarado, Verspoor, and Baldwin}]{alvarado2015domain}
Julio Cesar~Salinas Alvarado, Karin Verspoor, and Timothy Baldwin. 2015.
\newblock Domain adaption of named entity recognition to support credit risk assessment.
\newblock In \emph{Proceedings of the Australasian Language Technology Association Workshop 2015}, pages 84--90.

\bibitem[{Athan et~al.(2013)Athan, Boley, Governatori, Palmirani, Paschke, and Wyner}]{athan2013oasis}
Tara Athan, Harold Boley, Guido Governatori, Monica Palmirani, Adrian Paschke, and Adam Wyner. 2013.
\newblock Oasis legalruleml.
\newblock In \emph{proceedings of the fourteenth international conference on artificial intelligence and law}, pages 3--12.

\bibitem[{Balas et~al.(2024)Balas, Wadden, H{\'e}bert, Mathison, Warren, Seavilleklein, Wyzynski, Callahan, Crawford, Arjmand et~al.}]{balas2024exploring}
Michael Balas, Jordan~Joseph Wadden, Philip~C H{\'e}bert, Eric Mathison, Marika~D Warren, Victoria Seavilleklein, Daniel Wyzynski, Alison Callahan, Sean~A Crawford, Parnian Arjmand, et~al. 2024.
\newblock Exploring the potential utility of ai large language models for medical ethics: an expert panel evaluation of gpt-4.
\newblock \emph{Journal of Medical Ethics}, 50(2):90--96.

\bibitem[{Banerjee and Lavie(2005)}]{banerjee2005meteor}
Satanjeev Banerjee and Alon Lavie. 2005.
\newblock Meteor: An automatic metric for mt evaluation with improved correlation with human judgments.
\newblock In \emph{Proceedings of the acl workshop on intrinsic and extrinsic evaluation measures for machine translation and/or summarization}, pages 65--72.

\bibitem[{Bang et~al.(2023)Bang, Cahyawijaya, Lee, Dai, Su, Wilie, Lovenia, Ji, Yu, Chung, Do, Xu, and Fung}]{bang2023multitask}
Yejin Bang, Samuel Cahyawijaya, Nayeon Lee, Wenliang Dai, Dan Su, Bryan Wilie, Holy Lovenia, Ziwei Ji, Tiezheng Yu, Willy Chung, Quyet~V. Do, Yan Xu, and Pascale Fung. 2023.
\newblock \href {http://arxiv.org/abs/2302.04023} {A multitask, multilingual, multimodal evaluation of chatgpt on reasoning, hallucination, and interactivity}.

\bibitem[{Benoit(2023)}]{benoit2023chatgpt}
James~RA Benoit. 2023.
\newblock Chatgpt for clinical vignette generation, revision, and evaluation.
\newblock \emph{MedRxiv}, pages 2023--02.

\bibitem[{Bhagavatula et~al.(2019)Bhagavatula, Bras, Malaviya, Sakaguchi, Holtzman, Rashkin, Downey, Yih, and Choi}]{bhagavatula2019abductive}
Chandra Bhagavatula, Ronan~Le Bras, Chaitanya Malaviya, Keisuke Sakaguchi, Ari Holtzman, Hannah Rashkin, Doug Downey, Scott Wen-tau Yih, and Yejin Choi. 2019.
\newblock Abductive commonsense reasoning.
\newblock \emph{arXiv preprint arXiv:1908.05739}.

\bibitem[{Bian et~al.(2024)Bian, Han, Sun, Lin, Lu, He, Jiang, and Dong}]{bian2024chatgpt}
Ning Bian, Xianpei Han, Le~Sun, Hongyu Lin, Yaojie Lu, Ben He, Shanshan Jiang, and Bin Dong. 2024.
\newblock \href {http://arxiv.org/abs/2303.16421} {Chatgpt is a knowledgeable but inexperienced solver: An investigation of commonsense problem in large language models}.

\bibitem[{Bisk et~al.(2020)Bisk, Zellers, Gao, Choi et~al.}]{bisk2020piqa}
Yonatan Bisk, Rowan Zellers, Jianfeng Gao, Yejin Choi, et~al. 2020.
\newblock Piqa: Reasoning about physical commonsense in natural language.
\newblock In \emph{Proceedings of the AAAI conference on artificial intelligence}, volume~34, pages 7432--7439.

\bibitem[{Blair-Stanek et~al.(2023)Blair-Stanek, Holzenberger, and Van~Durme}]{blair2023can}
Andrew Blair-Stanek, Nils Holzenberger, and Benjamin Van~Durme. 2023.
\newblock Can gpt-3 perform statutory reasoning?
\newblock In \emph{Proceedings of the Nineteenth International Conference on Artificial Intelligence and Law}, pages 22--31.

\bibitem[{Cai et~al.(2024)Cai, Wang, Ma, Chen, and Zhou}]{cai2024large}
Tianle Cai, Xuezhi Wang, Tengyu Ma, Xinyun Chen, and Denny Zhou. 2024.
\newblock \href {http://arxiv.org/abs/2305.17126} {Large language models as tool makers}.

\bibitem[{Carlini et~al.(2023)Carlini, Ippolito, Jagielski, Lee, Tramer, and Zhang}]{carlini2023quantifying}
Nicholas Carlini, Daphne Ippolito, Matthew Jagielski, Katherine Lee, Florian Tramer, and Chiyuan Zhang. 2023.
\newblock \href {http://arxiv.org/abs/2202.07646} {Quantifying memorization across neural language models}.

\bibitem[{Casino et~al.(2022)Casino, Dasaklis, Spathoulas, Anagnostopoulos, Ghosal, Borocz, Solanas, Conti, and Patsakis}]{casino2022research}
Fran Casino, Thomas~K Dasaklis, Georgios~P Spathoulas, Marios Anagnostopoulos, Amrita Ghosal, Istvan Borocz, Agusti Solanas, Mauro Conti, and Constantinos Patsakis. 2022.
\newblock Research trends, challenges, and emerging topics in digital forensics: A review of reviews.
\newblock \emph{IEEE Access}, 10:25464--25493.

\bibitem[{Chang et~al.(2023)Chang, Wang, Wang, Wu, Yang, Zhu, Chen, Yi, Wang, Wang et~al.}]{chang2023survey}
Yupeng Chang, Xu~Wang, Jindong Wang, Yuan Wu, Linyi Yang, Kaijie Zhu, Hao Chen, Xiaoyuan Yi, Cunxiang Wang, Yidong Wang, et~al. 2023.
\newblock A survey on evaluation of large language models.
\newblock \emph{ACM Transactions on Intelligent Systems and Technology}.

\bibitem[{Chen et~al.(2020)Chen, Zha, Chen, Xiong, Wang, and Wang}]{chen2020hybridqa}
Wenhu Chen, Hanwen Zha, Zhiyu Chen, Wenhan Xiong, Hong Wang, and William Wang. 2020.
\newblock Hybridqa: A dataset of multi-hop question answering over tabular and textual data.
\newblock \emph{arXiv preprint arXiv:2004.07347}.

\bibitem[{Chen et~al.(2022)Chen, Li, Smiley, Ma, Shah, and Wang}]{chen2022convfinqa}
Zhiyu Chen, Shiyang Li, Charese Smiley, Zhiqiang Ma, Sameena Shah, and William~Yang Wang. 2022.
\newblock Convfinqa: Exploring the chain of numerical reasoning in conversational finance question answering.
\newblock \emph{arXiv preprint arXiv:2210.03849}.

\bibitem[{Cifuentes et~al.(2022)Cifuentes, Sandoval~Orozco, and Garcia~Villalba}]{cifuentes2022survey}
Jenny Cifuentes, Ana~Lucila Sandoval~Orozco, and Luis~Javier Garcia~Villalba. 2022.
\newblock A survey of artificial intelligence strategies for automatic detection of sexually explicit videos.
\newblock \emph{Multimedia Tools and Applications}, 81(3):3205--3222.

\bibitem[{Cobbe et~al.(2021)Cobbe, Kosaraju, Bavarian, Chen, Jun, Kaiser, Plappert, Tworek, Hilton, Nakano et~al.}]{cobbe2021training}
Karl Cobbe, Vineet Kosaraju, Mohammad Bavarian, Mark Chen, Heewoo Jun, Lukasz Kaiser, Matthias Plappert, Jerry Tworek, Jacob Hilton, Reiichiro Nakano, et~al. 2021.
\newblock Training verifiers to solve math word problems.
\newblock \emph{arXiv preprint arXiv:2110.14168}.

\bibitem[{Dalvi et~al.(2021)Dalvi, Jansen, Tafjord, Xie, Smith, Pipatanangkura, and Clark}]{dalvi2021explaining}
Bhavana Dalvi, Peter Jansen, Oyvind Tafjord, Zhengnan Xie, Hannah Smith, Leighanna Pipatanangkura, and Peter Clark. 2021.
\newblock Explaining answers with entailment trees.
\newblock \emph{arXiv preprint arXiv:2104.08661}.

\bibitem[{Daniel et~al.(2008)Daniel, Sornette, and Wohrmann}]{daniel2008look}
Gilles Daniel, Didier Sornette, and Peter Wohrmann. 2008.
\newblock Look-ahead benchmark bias in portfolio performance evaluation.
\newblock \emph{arXiv preprint arXiv:0810.1922}.

\bibitem[{Das et~al.(2024)Das, Amini, and Wu}]{das2024security}
Badhan~Chandra Das, M~Hadi Amini, and Yanzhao Wu. 2024.
\newblock Security and privacy challenges of large language models: A survey.
\newblock \emph{arXiv preprint arXiv:2402.00888}.

\bibitem[{Demszky et~al.(2023)Demszky, Yang, Yeager, Bryan, Clapper, Chandhok, Eichstaedt, Hecht, Jamieson, Johnson et~al.}]{demszky2023using}
Dorottya Demszky, Diyi Yang, David~S Yeager, Christopher~J Bryan, Margarett Clapper, Susannah Chandhok, Johannes~C Eichstaedt, Cameron Hecht, Jeremy Jamieson, Meghann Johnson, et~al. 2023.
\newblock Using large language models in psychology.
\newblock \emph{Nature Reviews Psychology}, 2(11):688--701.

\bibitem[{Deng et~al.(2020)Deng, Awadallah, Meek, Polozov, Sun, and Richardson}]{deng2020structure}
Xiang Deng, Ahmed~Hassan Awadallah, Christopher Meek, Oleksandr Polozov, Huan Sun, and Matthew Richardson. 2020.
\newblock Structure-grounded pretraining for text-to-sql.
\newblock \emph{arXiv preprint arXiv:2010.12773}.

\bibitem[{Deng et~al.(2023)Deng, Gu, Zheng, Chen, Stevens, Wang, Sun, and Su}]{deng2023mind2web}
Xiang Deng, Yu~Gu, Boyuan Zheng, Shijie Chen, Samuel Stevens, Boshi Wang, Huan Sun, and Yu~Su. 2023.
\newblock \href {http://arxiv.org/abs/2306.06070} {Mind2web: Towards a generalist agent for the web}.

\bibitem[{Deroy et~al.(2023)Deroy, Ghosh, and Ghosh}]{deroy2023ready}
Aniket Deroy, Kripabandhu Ghosh, and Saptarshi Ghosh. 2023.
\newblock How ready are pre-trained abstractive models and llms for legal case judgement summarization?
\newblock \emph{arXiv preprint arXiv:2306.01248}.

\bibitem[{Dijkstra et~al.(2022)Dijkstra, Gen{\c{c}}, Kayal, Kamps et~al.}]{dijkstra2022reading}
Ramon Dijkstra, Z{\"u}lk{\"u}f Gen{\c{c}}, Subhradeep Kayal, Jaap Kamps, et~al. 2022.
\newblock Reading comprehension quiz generation using generative pre-trained transformers.
\newblock In \emph{iTextbooks@ AIED}, pages 4--17.

\bibitem[{Du et~al.(2014)Du, Tao, and Martinez}]{du2014compound}
Shichuan Du, Yong Tao, and Aleix~M Martinez. 2014.
\newblock Compound facial expressions of emotion.
\newblock \emph{Proceedings of the national academy of sciences}, 111(15):E1454--E1462.

\bibitem[{Duan et~al.(2024)Duan, Yi, Zhang, Lu, Xie, and Gu}]{duan2024denevil}
Shitong Duan, Xiaoyuan Yi, Peng Zhang, Tun Lu, Xing Xie, and Ning Gu. 2024.
\newblock \href {http://arxiv.org/abs/2310.11053} {Denevil: Towards deciphering and navigating the ethical values of large language models via instruction learning}.

\bibitem[{Dubois et~al.(2024)Dubois, Li, Taori, Zhang, Gulrajani, Ba, Guestrin, Liang, and Hashimoto}]{dubois2024alpacafarm}
Yann Dubois, Chen~Xuechen Li, Rohan Taori, Tianyi Zhang, Ishaan Gulrajani, Jimmy Ba, Carlos Guestrin, Percy~S Liang, and Tatsunori~B Hashimoto. 2024.
\newblock Alpacafarm: A simulation framework for methods that learn from human feedback.
\newblock \emph{Advances in Neural Information Processing Systems}, 36.

\bibitem[{Engel and Mcadams(2024)}]{engel2024asking}
Christoph Engel and Richard~H Mcadams. 2024.
\newblock Asking gpt for the ordinary meaning of statutory terms.
\newblock \emph{MPI Collective Goods Discussion Paper}, (2024/5).

\bibitem[{Fan et~al.(2019)Fan, Jernite, Perez, Grangier, Weston, and Auli}]{fan2019eli5}
Angela Fan, Yacine Jernite, Ethan Perez, David Grangier, Jason Weston, and Michael Auli. 2019.
\newblock Eli5: Long form question answering.
\newblock \emph{arXiv preprint arXiv:1907.09190}.

\bibitem[{Feldman et~al.(2023)Feldman, Foulds, and Pan}]{feldman2023trapping}
Philip Feldman, James~R. Foulds, and Shimei Pan. 2023.
\newblock \href {http://arxiv.org/abs/2306.06085} {Trapping llm hallucinations using tagged context prompts}.

\bibitem[{Gao et~al.(2023)Gao, Wang, Li, Sun, Qian, Ding, and Zhou}]{gao2023text}
Dawei Gao, Haibin Wang, Yaliang Li, Xiuyu Sun, Yichen Qian, Bolin Ding, and Jingren Zhou. 2023.
\newblock Text-to-sql empowered by large language models: A benchmark evaluation.
\newblock \emph{arXiv preprint arXiv:2308.15363}.

\bibitem[{Geva et~al.(2021)Geva, Khashabi, Segal, Khot, Roth, and Berant}]{geva2021did}
Mor Geva, Daniel Khashabi, Elad Segal, Tushar Khot, Dan Roth, and Jonathan Berant. 2021.
\newblock Did aristotle use a laptop? a question answering benchmark with implicit reasoning strategies.
\newblock \emph{Transactions of the Association for Computational Linguistics}, 9:346--361.

\bibitem[{Ghallab et~al.(2004)Ghallab, Nau, and Traverso}]{ghallab2004automated}
Malik Ghallab, Dana Nau, and Paolo Traverso. 2004.
\newblock \emph{Automated Planning: theory and practice}.
\newblock Elsevier.

\bibitem[{Ghosh et~al.(2024)Ghosh, Varshney, Galinkin, and Parisien}]{ghosh2024aegis}
Shaona Ghosh, Prasoon Varshney, Erick Galinkin, and Christopher Parisien. 2024.
\newblock \href {http://arxiv.org/abs/2404.05993} {Aegis: Online adaptive ai content safety moderation with ensemble of llm experts}.

\bibitem[{Guo et~al.(2023)Guo, Jin, Liu, Huang, Shi, Yu, Liu, Li, Xiong, Xiong et~al.}]{guo2023evaluating}
Zishan Guo, Renren Jin, Chuang Liu, Yufei Huang, Dan Shi, Linhao Yu, Yan Liu, Jiaxuan Li, Bojian Xiong, Deyi Xiong, et~al. 2023.
\newblock Evaluating large language models: A comprehensive survey.
\newblock \emph{arXiv preprint arXiv:2310.19736}.

\bibitem[{Han et~al.(2023)Han, Ransom, Perfors, and Kemp}]{han2023inductive}
Simon~J. Han, Keith Ransom, Andrew Perfors, and Charles Kemp. 2023.
\newblock \href {http://arxiv.org/abs/2306.06548} {Inductive reasoning in humans and large language models}.

\bibitem[{Hendrycks et~al.(2021)Hendrycks, Burns, Kadavath, Arora, Basart, Tang, Song, and Steinhardt}]{hendrycks2021measuring}
Dan Hendrycks, Collin Burns, Saurav Kadavath, Akul Arora, Steven Basart, Eric Tang, Dawn Song, and Jacob Steinhardt. 2021.
\newblock Measuring mathematical problem solving with the math dataset.
\newblock \emph{arXiv preprint arXiv:2103.03874}.

\bibitem[{Holzenberger et~al.(2020)Holzenberger, Blair-Stanek, and Van~Durme}]{holzenberger2020dataset}
Nils Holzenberger, Andrew Blair-Stanek, and Benjamin Van~Durme. 2020.
\newblock A dataset for statutory reasoning in tax law entailment and question answering.
\newblock \emph{arXiv preprint arXiv:2005.05257}.

\bibitem[{Hort et~al.(2021)Hort, Zhang, Sarro, and Harman}]{hort2021fairea}
Max Hort, Jie~M Zhang, Federica Sarro, and Mark Harman. 2021.
\newblock Fairea: A model behaviour mutation approach to benchmarking bias mitigation methods.
\newblock In \emph{Proceedings of the 29th ACM joint meeting on european software engineering conference and symposium on the foundations of software engineering}, pages 994--1006.

\bibitem[{Hu et~al.(2023)Hu, Fu, Du, Luo, Zhao, and Zhao}]{hu2023chatdb}
Chenxu Hu, Jie Fu, Chenzhuang Du, Simian Luo, Junbo Zhao, and Hang Zhao. 2023.
\newblock \href {http://arxiv.org/abs/2306.03901} {Chatdb: Augmenting llms with databases as their symbolic memory}.

\bibitem[{Huang and Chang(2023)}]{huang2023reasoning}
Jie Huang and Kevin Chen-Chuan Chang. 2023.
\newblock \href {http://arxiv.org/abs/2212.10403} {Towards reasoning in large language models: A survey}.

\bibitem[{Huang et~al.(2022{\natexlab{a}})Huang, Abbeel, Pathak, and Mordatch}]{huang2022language}
Wenlong Huang, Pieter Abbeel, Deepak Pathak, and Igor Mordatch. 2022{\natexlab{a}}.
\newblock Language models as zero-shot planners: Extracting actionable knowledge for embodied agents.
\newblock In \emph{International Conference on Machine Learning}, pages 9118--9147. PMLR.

\bibitem[{Huang et~al.(2023)Huang, Wang, Zhang, Li, Wu, and Fei-Fei}]{huang2023voxposer}
Wenlong Huang, Chen Wang, Ruohan Zhang, Yunzhu Li, Jiajun Wu, and Li~Fei-Fei. 2023.
\newblock Voxposer: Composable 3d value maps for robotic manipulation with language models.
\newblock \emph{arXiv preprint arXiv:2307.05973}.

\bibitem[{Huang et~al.(2022{\natexlab{b}})Huang, Xia, Xiao, Chan, Liang, Florence, Zeng, Tompson, Mordatch, Chebotar, Sermanet, Brown, Jackson, Luu, Levine, Hausman, and Ichter}]{huang2022inner}
Wenlong Huang, Fei Xia, Ted Xiao, Harris Chan, Jacky Liang, Pete Florence, Andy Zeng, Jonathan Tompson, Igor Mordatch, Yevgen Chebotar, Pierre Sermanet, Noah Brown, Tomas Jackson, Linda Luu, Sergey Levine, Karol Hausman, and Brian Ichter. 2022{\natexlab{b}}.
\newblock Inner monologue: Embodied reasoning through planning with language models.
\newblock In \emph{arXiv preprint arXiv:2207.05608}.

\bibitem[{Ji et~al.(2023)Ji, Liu, Dai, Pan, Zhang, Bian, Zhang, Sun, Wang, and Yang}]{ji2023beavertails}
Jiaming Ji, Mickel Liu, Juntao Dai, Xuehai Pan, Chi Zhang, Ce~Bian, Chi Zhang, Ruiyang Sun, Yizhou Wang, and Yaodong Yang. 2023.
\newblock \href {http://arxiv.org/abs/2307.04657} {Beavertails: Towards improved safety alignment of llm via a human-preference dataset}.

\bibitem[{Jia et~al.(2021)Jia, Cui, Xiao, Liu, Rashid, and Gehringer}]{jia2021all}
Qinjin Jia, Jialin Cui, Yunkai Xiao, Chengyuan Liu, Parvez Rashid, and Edward~F Gehringer. 2021.
\newblock All-in-one: Multi-task learning bert models for evaluating peer assessments.
\newblock \emph{arXiv preprint arXiv:2110.03895}.

\bibitem[{Jiang et~al.(2024)Jiang, Ye, Dong, Jia, Xu, Yan, Zhang, and Zhang}]{jiang2024hal}
Chaoya Jiang, Wei Ye, Mengfan Dong, Hongrui Jia, Haiyang Xu, Ming Yan, Ji~Zhang, and Shikun Zhang. 2024.
\newblock Hal-eval: A universal and fine-grained hallucination evaluation framework for large vision language models.
\newblock \emph{arXiv preprint arXiv:2402.15721}.

\bibitem[{Jiang et~al.(2020)Jiang, Bordia, Zhong, Dognin, Singh, and Bansal}]{jiang2020hover}
Yichen Jiang, Shikha Bordia, Zheng Zhong, Charles Dognin, Maneesh Singh, and Mohit Bansal. 2020.
\newblock \href {http://arxiv.org/abs/2011.03088} {Hover: A dataset for many-hop fact extraction and claim verification}.

\bibitem[{Jin et~al.(2024)Jin, Zhu, Wang, Zhou, Zhang, Zhang et~al.}]{jin2024attackeval}
Mingyu Jin, Suiyuan Zhu, Beichen Wang, Zihao Zhou, Chong Zhang, Yongfeng Zhang, et~al. 2024.
\newblock Attackeval: How to evaluate the effectiveness of jailbreak attacking on large language models.
\newblock \emph{arXiv preprint arXiv:2401.09002}.

\bibitem[{Karamizadeh et~al.(2023)Karamizadeh, Shojae~Chaeikar, and Jolfaei}]{karamizadeh2023adult}
Sasan Karamizadeh, Saman Shojae~Chaeikar, and Alireza Jolfaei. 2023.
\newblock Adult content image recognition by boltzmann machine limited and deep learning.
\newblock \emph{Evolutionary Intelligence}, 16(4):1185--1194.

\bibitem[{Karinshak et~al.(2023)Karinshak, Liu, Park, and Hancock}]{karinshak2023working}
Elise Karinshak, Sunny~Xun Liu, Joon~Sung Park, and Jeffrey~T Hancock. 2023.
\newblock Working with ai to persuade: Examining a large language model's ability to generate pro-vaccination messages.
\newblock \emph{Proceedings of the ACM on Human-Computer Interaction}, 7(CSCW1):1--29.

\bibitem[{Kasneci et~al.(2023)Kasneci, Se{\ss}ler, K{\"u}chemann, Bannert, Dementieva, Fischer, Gasser, Groh, G{\"u}nnemann, H{\"u}llermeier et~al.}]{kasneci2023chatgpt}
Enkelejda Kasneci, Kathrin Se{\ss}ler, Stefan K{\"u}chemann, Maria Bannert, Daryna Dementieva, Frank Fischer, Urs Gasser, Georg Groh, Stephan G{\"u}nnemann, Eyke H{\"u}llermeier, et~al. 2023.
\newblock Chatgpt for good? on opportunities and challenges of large language models for education.
\newblock \emph{Learning and individual differences}, 103:102274.

\bibitem[{Katz et~al.(2024)Katz, Bommarito, Gao, and Arredondo}]{katz2024gpt}
Daniel~Martin Katz, Michael~James Bommarito, Shang Gao, and Pablo Arredondo. 2024.
\newblock Gpt-4 passes the bar exam.
\newblock \emph{Philosophical Transactions of the Royal Society A}, 382(2270):20230254.

\bibitem[{Khan et~al.(2024)Khan, Hughes, Valentine, Ruis, Sachan, Radhakrishnan, Grefenstette, Bowman, Rockt{\"a}schel, and Perez}]{khan2024debating}
Akbir Khan, John Hughes, Dan Valentine, Laura Ruis, Kshitij Sachan, Ansh Radhakrishnan, Edward Grefenstette, Samuel~R Bowman, Tim Rockt{\"a}schel, and Ethan Perez. 2024.
\newblock Debating with more persuasive llms leads to more truthful answers.
\newblock \emph{arXiv preprint arXiv:2402.06782}.

\bibitem[{Kim et~al.(2024{\natexlab{a}})Kim, Yeen, and Koo}]{kim2024towards}
Minju Kim, Heuiyeen Yeen, and Myoung-Wan Koo. 2024{\natexlab{a}}.
\newblock Towards context-based violence detection: A korean crime dialogue dataset.
\newblock In \emph{Findings of the Association for Computational Linguistics: EACL 2024}, pages 603--623.

\bibitem[{Kim et~al.(2024{\natexlab{b}})Kim, Yun, Lee, Gubri, Yoon, and Oh}]{kim2024propile}
Siwon Kim, Sangdoo Yun, Hwaran Lee, Martin Gubri, Sungroh Yoon, and Seong~Joon Oh. 2024{\natexlab{b}}.
\newblock Propile: Probing privacy leakage in large language models.
\newblock \emph{Advances in Neural Information Processing Systems}, 36.

\bibitem[{Kosinski(2023)}]{kosinski2023theory}
Michal Kosinski. 2023.
\newblock Theory of mind may have spontaneously emerged in large language models.
\newblock \emph{arXiv preprint arXiv:2302.02083}, 4:169.

\bibitem[{Kumar(2023)}]{kumar2023analysis}
Arun~HS Kumar. 2023.
\newblock Analysis of chatgpt tool to assess the potential of its utility for academic writing in biomedical domain.
\newblock \emph{Biology, Engineering, Medicine and Science Reports}, 9(1):24--30.

\bibitem[{Kung et~al.(2023)Kung, Cheatham, Medenilla, Sillos, De~Leon, Elepa{\~n}o, Madriaga, Aggabao, Diaz-Candido, Maningo et~al.}]{kung2023performance}
Tiffany~H Kung, Morgan Cheatham, Arielle Medenilla, Czarina Sillos, Lorie De~Leon, Camille Elepa{\~n}o, Maria Madriaga, Rimel Aggabao, Giezel Diaz-Candido, James Maningo, et~al. 2023.
\newblock Performance of chatgpt on usmle: potential for ai-assisted medical education using large language models.
\newblock \emph{PLoS digital health}, 2(2):e0000198.

\bibitem[{Kwiatkowski et~al.(2019)Kwiatkowski, Palomaki, Redfield, Collins, Parikh, Alberti, Epstein, Polosukhin, Devlin, Lee et~al.}]{kwiatkowski2019natural}
Tom Kwiatkowski, Jennimaria Palomaki, Olivia Redfield, Michael Collins, Ankur Parikh, Chris Alberti, Danielle Epstein, Illia Polosukhin, Jacob Devlin, Kenton Lee, et~al. 2019.
\newblock Natural questions: a benchmark for question answering research.
\newblock \emph{Transactions of the Association for Computational Linguistics}, 7:453--466.

\bibitem[{Lee et~al.(2024)Lee, Stevens, Han, and Song}]{lee2024survey}
Jean Lee, Nicholas Stevens, Soyeon~Caren Han, and Minseok Song. 2024.
\newblock A survey of large language models in finance (finllms).
\newblock \emph{arXiv preprint arXiv:2402.02315}.

\bibitem[{Li et~al.(2024{\natexlab{a}})Li, Dong, Wang, Hu, Zuo, Lin, Qiao, and Shao}]{li2024saladbench}
Lijun Li, Bowen Dong, Ruohui Wang, Xuhao Hu, Wangmeng Zuo, Dahua Lin, Yu~Qiao, and Jing Shao. 2024{\natexlab{a}}.
\newblock \href {http://arxiv.org/abs/2402.05044} {Salad-bench: A hierarchical and comprehensive safety benchmark for large language models}.

\bibitem[{Li et~al.(2023{\natexlab{a}})Li, Zhao, Yu, Song, Li, Yu, Li, Huang, and Li}]{li2023apibank}
Minghao Li, Yingxiu Zhao, Bowen Yu, Feifan Song, Hangyu Li, Haiyang Yu, Zhoujun Li, Fei Huang, and Yongbin Li. 2023{\natexlab{a}}.
\newblock \href {http://arxiv.org/abs/2304.08244} {Api-bank: A comprehensive benchmark for tool-augmented llms}.

\bibitem[{Li et~al.(2023{\natexlab{b}})Li, Wang, Ding, and Chen}]{li2023large}
Yinheng Li, Shaofei Wang, Han Ding, and Hang Chen. 2023{\natexlab{b}}.
\newblock Large language models in finance: A survey.
\newblock In \emph{Proceedings of the Fourth ACM International Conference on AI in Finance}, pages 374--382.

\bibitem[{Li et~al.(2024{\natexlab{b}})Li, Xu, Shen, Xu, Gu, and Tao}]{li2024leveraging}
Zhen Li, Xiaohan Xu, Tao Shen, Can Xu, Jia-Chen Gu, and Chongyang Tao. 2024{\natexlab{b}}.
\newblock Leveraging large language models for nlg evaluation: A survey.
\newblock \emph{arXiv preprint arXiv:2401.07103}.

\bibitem[{Liang et~al.(2023)Liang, Huang, Xia, Xu, Hausman, Ichter, Florence, and Zeng}]{liang2023code}
Jacky Liang, Wenlong Huang, Fei Xia, Peng Xu, Karol Hausman, Brian Ichter, Pete Florence, and Andy Zeng. 2023.
\newblock Code as policies: Language model programs for embodied control.
\newblock In \emph{2023 IEEE International Conference on Robotics and Automation (ICRA)}, pages 9493--9500. IEEE.

\bibitem[{Liang et~al.(2022)Liang, Bommasani, Lee, Tsipras, Soylu, Yasunaga, Zhang, Narayanan, Wu, Kumar et~al.}]{liang2022holistic}
Percy Liang, Rishi Bommasani, Tony Lee, Dimitris Tsipras, Dilara Soylu, Michihiro Yasunaga, Yian Zhang, Deepak Narayanan, Yuhuai Wu, Ananya Kumar, et~al. 2022.
\newblock Holistic evaluation of language models.
\newblock \emph{arXiv preprint arXiv:2211.09110}.

\bibitem[{Liang et~al.(2019)Liang, Li, and Yin}]{liang2019new}
Yichan Liang, Jianheng Li, and Jian Yin. 2019.
\newblock A new multi-choice reading comprehension dataset for curriculum learning.
\newblock In \emph{Asian Conference on Machine Learning}, pages 742--757. PMLR.

\bibitem[{Liga and Robaldo(2023)}]{liga2023fine}
Davide Liga and Livio Robaldo. 2023.
\newblock Fine-tuning gpt-3 for legal rule classification.
\newblock \emph{Computer Law \& Security Review}, 51:105864.

\bibitem[{Lin(2004)}]{lin2004rouge}
Chin-Yew Lin. 2004.
\newblock Rouge: A package for automatic evaluation of summaries.
\newblock In \emph{Text summarization branches out}, pages 74--81.

\bibitem[{Lin et~al.(2022)Lin, Hilton, and Evans}]{lin2022truthfulqa}
Stephanie Lin, Jacob Hilton, and Owain Evans. 2022.
\newblock \href {http://arxiv.org/abs/2109.07958} {Truthfulqa: Measuring how models mimic human falsehoods}.

\bibitem[{Lin et~al.(2023)Lin, Wang, Tong, Wang, Guo, Wang, and Shang}]{lin2023toxicchat}
Zi~Lin, Zihan Wang, Yongqi Tong, Yangkun Wang, Yuxin Guo, Yujia Wang, and Jingbo Shang. 2023.
\newblock \href {http://arxiv.org/abs/2310.17389} {Toxicchat: Unveiling hidden challenges of toxicity detection in real-world user-ai conversation}.

\bibitem[{Liu et~al.(2024{\natexlab{a}})Liu, Liu, Shi, Huang, Wang, Yang, and Zhang}]{liu2024exploring}
Fang Liu, Yang Liu, Lin Shi, Houkun Huang, Ruifeng Wang, Zhen Yang, and Li~Zhang. 2024{\natexlab{a}}.
\newblock Exploring and evaluating hallucinations in llm-powered code generation.
\newblock \emph{arXiv preprint arXiv:2404.00971}.

\bibitem[{Liu et~al.(2023)Liu, Ning, Teng, Liu, Zhou, and Zhang}]{liu2023evaluating}
Hanmeng Liu, Ruoxi Ning, Zhiyang Teng, Jian Liu, Qiji Zhou, and Yue Zhang. 2023.
\newblock \href {http://arxiv.org/abs/2304.03439} {Evaluating the logical reasoning ability of chatgpt and gpt-4}.

\bibitem[{Liu et~al.(2024{\natexlab{b}})Liu, Yu, Zhang, Zhang, and Xiao}]{liu2024automatic}
Xiaogeng Liu, Zhiyuan Yu, Yizhe Zhang, Ning Zhang, and Chaowei Xiao. 2024{\natexlab{b}}.
\newblock Automatic and universal prompt injection attacks against large language models.
\newblock \emph{arXiv preprint arXiv:2403.04957}.

\bibitem[{Liu et~al.(2021{\natexlab{a}})Liu, Shi, Chen, Yu, Li, and Zhao}]{liu2021imigue}
Xin Liu, Henglin Shi, Haoyu Chen, Zitong Yu, Xiaobai Li, and Guoying Zhao. 2021{\natexlab{a}}.
\newblock imigue: An identity-free video dataset for micro-gesture understanding and emotion analysis.
\newblock In \emph{Proceedings of the IEEE/CVF conference on computer vision and pattern recognition}, pages 10631--10642.

\bibitem[{Liu et~al.(2021{\natexlab{b}})Liu, Huang, Huang, Li, and Zhao}]{liu2021finbert}
Zhuang Liu, Degen Huang, Kaiyu Huang, Zhuang Li, and Jun Zhao. 2021{\natexlab{b}}.
\newblock Finbert: A pre-trained financial language representation model for financial text mining.
\newblock In \emph{Proceedings of the twenty-ninth international conference on international joint conferences on artificial intelligence}, pages 4513--4519.

\bibitem[{Lu et~al.(2024)Lu, Niu, Wang, Wang, Hu, Tang, Zhang, Yuan, Huang, Yu et~al.}]{lu2024gpt}
Hao Lu, Xuesong Niu, Jiyao Wang, Yin Wang, Qingyong Hu, Jiaqi Tang, Yuting Zhang, Kaishen Yuan, Bin Huang, Zitong Yu, et~al. 2024.
\newblock Gpt as psychologist? preliminary evaluations for gpt-4v on visual affective computing.
\newblock \emph{arXiv preprint arXiv:2403.05916}.

\bibitem[{Mahowald et~al.(2023)Mahowald, Ivanova, Blank, Kanwisher, Tenenbaum, and Fedorenko}]{mahowald2023dissociating}
Kyle Mahowald, Anna~A Ivanova, Idan~A Blank, Nancy Kanwisher, Joshua~B Tenenbaum, and Evelina Fedorenko. 2023.
\newblock Dissociating language and thought in large language models: a cognitive perspective.
\newblock \emph{arXiv preprint arXiv:2301.06627}.

\bibitem[{Maia et~al.(2018)Maia, Handschuh, Freitas, Davis, McDermott, Zarrouk, and Balahur}]{maia201818}
Macedo Maia, Siegfried Handschuh, Andr{\'e} Freitas, Brian Davis, Ross McDermott, Manel Zarrouk, and Alexandra Balahur. 2018.
\newblock Www'18 open challenge: financial opinion mining and question answering.
\newblock In \emph{Companion proceedings of the the web conference 2018}, pages 1941--1942.

\bibitem[{Malo et~al.(2014)Malo, Sinha, Korhonen, Wallenius, and Takala}]{malo2014good}
Pekka Malo, Ankur Sinha, Pekka Korhonen, Jyrki Wallenius, and Pyry Takala. 2014.
\newblock Good debt or bad debt: Detecting semantic orientations in economic texts.
\newblock \emph{Journal of the Association for Information Science and Technology}, 65(4):782--796.

\bibitem[{Mann et~al.(2020)Mann, Ryder, Subbiah, Kaplan, Dhariwal, Neelakantan, Shyam, Sastry, Askell, Agarwal et~al.}]{mann2020language}
Ben Mann, N~Ryder, M~Subbiah, J~Kaplan, P~Dhariwal, A~Neelakantan, P~Shyam, G~Sastry, A~Askell, S~Agarwal, et~al. 2020.
\newblock Language models are few-shot learners.
\newblock \emph{arXiv preprint arXiv:2005.14165}.

\bibitem[{Markov et~al.(2023)Markov, Zhang, Agarwal, Eloundou, Lee, Adler, Jiang, and Weng}]{markov2023holistic}
Todor Markov, Chong Zhang, Sandhini Agarwal, Tyna Eloundou, Teddy Lee, Steven Adler, Angela Jiang, and Lilian Weng. 2023.
\newblock \href {http://arxiv.org/abs/2208.03274} {A holistic approach to undesired content detection in the real world}.

\bibitem[{Mavadati et~al.(2013)Mavadati, Mahoor, Bartlett, Trinh, and Cohn}]{mavadati2013disfa}
S~Mohammad Mavadati, Mohammad~H Mahoor, Kevin Bartlett, Philip Trinh, and Jeffrey~F Cohn. 2013.
\newblock Disfa: A spontaneous facial action intensity database.
\newblock \emph{IEEE Transactions on Affective Computing}, 4(2):151--160.

\bibitem[{Menick et~al.(2022)Menick, Trebacz, Mikulik, Aslanides, Song, Chadwick, Glaese, Young, Campbell-Gillingam, Irving et~al.}]{menick2022teaching}
Jacob Menick, Maja Trebacz, Vladimir Mikulik, John Aslanides, Francis Song, Martin Chadwick, Mia Glaese, Susannah Young, Lucy Campbell-Gillingam, Geoffrey Irving, et~al. 2022.
\newblock Teaching language models to support answers with verified quotes. arxiv.

\bibitem[{Miao et~al.(2023)Miao, Teh, and Rainforth}]{miao2023selfcheck}
Ning Miao, Yee~Whye Teh, and Tom Rainforth. 2023.
\newblock Selfcheck: Using llms to zero-shot check their own step-by-step reasoning.
\newblock \emph{arXiv preprint arXiv:2308.00436}.

\bibitem[{Mihaylov et~al.(2018)Mihaylov, Clark, Khot, and Sabharwal}]{OpenBookQA2018}
Todor Mihaylov, Peter Clark, Tushar Khot, and Ashish Sabharwal. 2018.
\newblock Can a suit of armor conduct electricity? a new dataset for open book question answering.
\newblock In \emph{EMNLP}.

\bibitem[{Min et~al.(2023)Min, Ross, Sulem, Veyseh, Nguyen, Sainz, Agirre, Heintz, and Roth}]{min2023recent}
Bonan Min, Hayley Ross, Elior Sulem, Amir Pouran~Ben Veyseh, Thien~Huu Nguyen, Oscar Sainz, Eneko Agirre, Ilana Heintz, and Dan Roth. 2023.
\newblock Recent advances in natural language processing via large pre-trained language models: A survey.
\newblock \emph{ACM Computing Surveys}, 56(2):1--40.

\bibitem[{Moon et~al.(2014)Moon, Pakhomov, Liu, Ryan, and Melton}]{moon2014sense}
Sungrim Moon, Serguei Pakhomov, Nathan Liu, James~O Ryan, and Genevieve~B Melton. 2014.
\newblock A sense inventory for clinical abbreviations and acronyms created using clinical notes and medical dictionary resources.
\newblock \emph{Journal of the American Medical Informatics Association}, 21(2):299--307.

\bibitem[{Muthukrishnan et~al.(2020)Muthukrishnan, Maleki, Ovens, Reinhold, Forghani, Forghani et~al.}]{muthukrishnan2020brief}
Nikesh Muthukrishnan, Farhad Maleki, Katie Ovens, Caroline Reinhold, Behzad Forghani, Reza Forghani, et~al. 2020.
\newblock Brief history of artificial intelligence.
\newblock \emph{Neuroimaging Clinics of North America}, 30(4):393--399.

\bibitem[{Nakano et~al.(2022)Nakano, Hilton, Balaji, Wu, Ouyang, Kim, Hesse, Jain, Kosaraju, Saunders, Jiang, Cobbe, Eloundou, Krueger, Button, Knight, Chess, and Schulman}]{nakano2022webgpt}
Reiichiro Nakano, Jacob Hilton, Suchir Balaji, Jeff Wu, Long Ouyang, Christina Kim, Christopher Hesse, Shantanu Jain, Vineet Kosaraju, William Saunders, Xu~Jiang, Karl Cobbe, Tyna Eloundou, Gretchen Krueger, Kevin Button, Matthew Knight, Benjamin Chess, and John Schulman. 2022.
\newblock \href {http://arxiv.org/abs/2112.09332} {Webgpt: Browser-assisted question-answering with human feedback}.

\bibitem[{Nayerifard et~al.(2023)Nayerifard, Amintoosi, Bafghi, and Dehghantanha}]{nayerifard2023machine}
Tahereh Nayerifard, Haleh Amintoosi, Abbas~Ghaemi Bafghi, and Ali Dehghantanha. 2023.
\newblock Machine learning in digital forensics: a systematic literature review.
\newblock \emph{arXiv preprint arXiv:2306.04965}.

\bibitem[{Nolfi(2023)}]{nolfi2023unexpected}
Stefano Nolfi. 2023.
\newblock On the unexpected abilities of large language models.
\newblock \emph{arXiv preprint arXiv:2308.09720}.

\bibitem[{Oviedo-Trespalacios et~al.(2023)Oviedo-Trespalacios, Peden, Cole-Hunter, Costantini, Haghani, Rod, Kelly, Torkamaan, Tariq, Newton et~al.}]{oviedo2023risks}
Oscar Oviedo-Trespalacios, Amy~E Peden, Thomas Cole-Hunter, Arianna Costantini, Milad Haghani, JE~Rod, Sage Kelly, Helma Torkamaan, Amina Tariq, James David~Albert Newton, et~al. 2023.
\newblock The risks of using chatgpt to obtain common safety-related information and advice.
\newblock \emph{Safety science}, 167:106244.

\bibitem[{Palmirani and Vitali(2011)}]{palmirani2011akoma}
Monica Palmirani and Fabio Vitali. 2011.
\newblock Akoma-ntoso for legal documents.
\newblock \emph{Legislative XML for the Semantic Web: Principles, Models, Standards for Document Management}, pages 75--100.

\bibitem[{Papineni et~al.(2002)Papineni, Roukos, Ward, and Zhu}]{papineni2002bleu}
Kishore Papineni, Salim Roukos, Todd Ward, and Wei-Jing Zhu. 2002.
\newblock Bleu: a method for automatic evaluation of machine translation.
\newblock In \emph{Proceedings of the 40th annual meeting of the Association for Computational Linguistics}, pages 311--318.

\bibitem[{Parrish et~al.(2021)Parrish, Chen, Nangia, Padmakumar, Phang, Thompson, Htut, and Bowman}]{parrish2021bbq}
Alicia Parrish, Angelica Chen, Nikita Nangia, Vishakh Padmakumar, Jason Phang, Jana Thompson, Phu~Mon Htut, and Samuel~R Bowman. 2021.
\newblock Bbq: A hand-built bias benchmark for question answering.
\newblock \emph{arXiv preprint arXiv:2110.08193}.

\bibitem[{Patil et~al.(2023)Patil, Zhang, Wang, and Gonzalez}]{patil2023gorilla}
Shishir~G. Patil, Tianjun Zhang, Xin Wang, and Joseph~E. Gonzalez. 2023.
\newblock \href {http://arxiv.org/abs/2305.15334} {Gorilla: Large language model connected with massive apis}.

\bibitem[{Petroni et~al.(2019)Petroni, Rockt{\"a}schel, Lewis, Bakhtin, Wu, Miller, and Riedel}]{petroni2019language}
Fabio Petroni, Tim Rockt{\"a}schel, Patrick Lewis, Anton Bakhtin, Yuxiang Wu, Alexander~H Miller, and Sebastian Riedel. 2019.
\newblock Language models as knowledge bases?
\newblock \emph{arXiv preprint arXiv:1909.01066}.

\bibitem[{Pinar~Saygin et~al.(2000)Pinar~Saygin, Cicekli, and Akman}]{pinar2000turing}
Ayse Pinar~Saygin, Ilyas Cicekli, and Varol Akman. 2000.
\newblock Turing test: 50 years later.
\newblock \emph{Minds and machines}, 10(4):463--518.

\bibitem[{Qian et~al.(2023)Qian, Han, Fung, Qin, Liu, and Ji}]{qian2023creator}
Cheng Qian, Chi Han, Yi~R. Fung, Yujia Qin, Zhiyuan Liu, and Heng Ji. 2023.
\newblock \href {http://arxiv.org/abs/2305.14318} {Creator: Tool creation for disentangling abstract and concrete reasoning of large language models}.

\bibitem[{Qin et~al.(2023{\natexlab{a}})Qin, Cai, Jin, Yan, Liang, Zhu, Lin, Han, Ding, Wang, Xie, Qi, Liu, Sun, and Zhou}]{qin2023webcpm}
Yujia Qin, Zihan Cai, Dian Jin, Lan Yan, Shihao Liang, Kunlun Zhu, Yankai Lin, Xu~Han, Ning Ding, Huadong Wang, Ruobing Xie, Fanchao Qi, Zhiyuan Liu, Maosong Sun, and Jie Zhou. 2023{\natexlab{a}}.
\newblock \href {http://arxiv.org/abs/2305.06849} {Webcpm: Interactive web search for chinese long-form question answering}.

\bibitem[{Qin et~al.(2023{\natexlab{b}})Qin, Hu, Lin, Chen, Ding, Cui, Zeng, Huang, Xiao, Han, Fung, Su, Wang, Qian, Tian, Zhu, Liang, Shen, Xu, Zhang, Ye, Li, Tang, Yi, Zhu, Dai, Yan, Cong, Lu, Zhao, Huang, Yan, Han, Sun, Li, Phang, Yang, Wu, Ji, Liu, and Sun}]{qin2023tool}
Yujia Qin, Shengding Hu, Yankai Lin, Weize Chen, Ning Ding, Ganqu Cui, Zheni Zeng, Yufei Huang, Chaojun Xiao, Chi Han, Yi~Ren Fung, Yusheng Su, Huadong Wang, Cheng Qian, Runchu Tian, Kunlun Zhu, Shihao Liang, Xingyu Shen, Bokai Xu, Zhen Zhang, Yining Ye, Bowen Li, Ziwei Tang, Jing Yi, Yuzhang Zhu, Zhenning Dai, Lan Yan, Xin Cong, Yaxi Lu, Weilin Zhao, Yuxiang Huang, Junxi Yan, Xu~Han, Xian Sun, Dahai Li, Jason Phang, Cheng Yang, Tongshuang Wu, Heng Ji, Zhiyuan Liu, and Maosong Sun. 2023{\natexlab{b}}.
\newblock \href {http://arxiv.org/abs/2304.08354} {Tool learning with foundation models}.

\bibitem[{Qin et~al.(2023{\natexlab{c}})Qin, Liang, Ye, Zhu, Yan, Lu, Lin, Cong, Tang, Qian, Zhao, Hong, Tian, Xie, Zhou, Gerstein, Li, Liu, and Sun}]{qin2023toolllm}
Yujia Qin, Shihao Liang, Yining Ye, Kunlun Zhu, Lan Yan, Yaxi Lu, Yankai Lin, Xin Cong, Xiangru Tang, Bill Qian, Sihan Zhao, Lauren Hong, Runchu Tian, Ruobing Xie, Jie Zhou, Mark Gerstein, Dahai Li, Zhiyuan Liu, and Maosong Sun. 2023{\natexlab{c}}.
\newblock \href {http://arxiv.org/abs/2307.16789} {Toolllm: Facilitating large language models to master 16000+ real-world apis}.

\bibitem[{Radford et~al.(2018)Radford, Narasimhan, Salimans, Sutskever et~al.}]{radford2018improving}
Alec Radford, Karthik Narasimhan, Tim Salimans, Ilya Sutskever, et~al. 2018.
\newblock Improving language understanding by generative pre-training.

\bibitem[{Radford et~al.(2019)Radford, Wu, Child, Luan, Amodei, Sutskever et~al.}]{radford2019language}
Alec Radford, Jeffrey Wu, Rewon Child, David Luan, Dario Amodei, Ilya Sutskever, et~al. 2019.
\newblock Language models are unsupervised multitask learners.
\newblock \emph{OpenAI blog}, 1(8):9.

\bibitem[{Raina and Gales(2022)}]{raina2022multiple}
Vatsal Raina and Mark Gales. 2022.
\newblock Multiple-choice question generation: Towards an automated assessment framework.
\newblock \emph{arXiv preprint arXiv:2209.11830}.

\bibitem[{Rana et~al.(2023)Rana, Haviland, Garg, Abou-Chakra, Reid, and Suenderhauf}]{rana2023sayplan}
Krishan Rana, Jesse Haviland, Sourav Garg, Jad Abou-Chakra, Ian Reid, and Niko Suenderhauf. 2023.
\newblock \href {https://openreview.net/forum?id=wMpOMO0Ss7a} {Sayplan: Grounding large language models using 3d scene graphs for scalable task planning}.
\newblock In \emph{7th Annual Conference on Robot Learning}.

\bibitem[{Rathje et~al.(2023)Rathje, Mirea, Sucholutsky, Marjieh, Robertson, and Van~Bavel}]{rathje2023gpt}
Steve Rathje, Dan-Mircea Mirea, Ilia Sucholutsky, Raja Marjieh, Claire Robertson, and Jay~J Van~Bavel. 2023.
\newblock Gpt is an effective tool for multilingual psychological text analysis.

\bibitem[{Ren et~al.(2023)Ren, Dixit, Bodrova, Singh, Tu, Brown, Xu, Takayama, Xia, Varley et~al.}]{ren2023robotaskforhelp}
Allen~Z Ren, Anushri Dixit, Alexandra Bodrova, Sumeet Singh, Stephen Tu, Noah Brown, Peng Xu, Leila Takayama, Fei Xia, Jake Varley, et~al. 2023.
\newblock Robots that ask for help: Uncertainty alignment for large language model planners.
\newblock \emph{arXiv preprint arXiv:2307.01928}.

\bibitem[{Ruan et~al.(2023)Ruan, Chen, Zhang, Xu, Bao, Du, Shi, Mao, Li, Zeng, and Zhao}]{ruan2023tptu}
Jingqing Ruan, Yihong Chen, Bin Zhang, Zhiwei Xu, Tianpeng Bao, Guoqing Du, Shiwei Shi, Hangyu Mao, Ziyue Li, Xingyu Zeng, and Rui Zhao. 2023.
\newblock \href {http://arxiv.org/abs/2308.03427} {Tptu: Large language model-based ai agents for task planning and tool usage}.

\bibitem[{Salehi and Burgue{\~n}o(2018)}]{salehi2018emerging}
Hadi Salehi and Rigoberto Burgue{\~n}o. 2018.
\newblock Emerging artificial intelligence methods in structural engineering.
\newblock \emph{Engineering structures}, 171:170--189.

\bibitem[{Scherrer et~al.(2023)Scherrer, Shi, Feder, and Blei}]{scherrer2023evaluating}
Nino Scherrer, Claudia Shi, Amir Feder, and David~M. Blei. 2023.
\newblock \href {http://arxiv.org/abs/2307.14324} {Evaluating the moral beliefs encoded in llms}.

\bibitem[{Schick et~al.(2023)Schick, Dwivedi-Yu, Dessì, Raileanu, Lomeli, Zettlemoyer, Cancedda, and Scialom}]{schick2023toolformer}
Timo Schick, Jane Dwivedi-Yu, Roberto Dessì, Roberta Raileanu, Maria Lomeli, Luke Zettlemoyer, Nicola Cancedda, and Thomas Scialom. 2023.
\newblock \href {http://arxiv.org/abs/2302.04761} {Toolformer: Language models can teach themselves to use tools}.

\bibitem[{Semigran et~al.(2015)Semigran, Linder, Gidengil, and Mehrotra}]{semigran2015evaluation}
Hannah~L Semigran, Jeffrey~A Linder, Courtney Gidengil, and Ateev Mehrotra. 2015.
\newblock Evaluation of symptom checkers for self diagnosis and triage: audit study.
\newblock \emph{bmj}, 351.

\bibitem[{Shah et~al.(2023)Shah, Equi, Osi{\'n}ski, Xia, Ichter, and Levine}]{shah2023heuristic_navigation}
Dhruv Shah, Michael~Robert Equi, B{\l}a{\.z}ej Osi{\'n}ski, Fei Xia, Brian Ichter, and Sergey Levine. 2023.
\newblock Navigation with large language models: Semantic guesswork as a heuristic for planning.
\newblock In \emph{Conference on Robot Learning}, pages 2683--2699. PMLR.

\bibitem[{Shah et~al.(2022)Shah, Osinski, Ichter, and Levine}]{shah2022lmnav}
Dhruv Shah, Blazej Osinski, Brian Ichter, and Sergey Levine. 2022.
\newblock \href {https://openreview.net/forum?id=UW5A3SweAH} {{LM}-nav: Robotic navigation with large pre-trained models of language, vision, and action}.
\newblock In \emph{6th Annual Conference on Robot Learning}.

\bibitem[{Shan and Deng(2018)}]{shan2018reliable}
Li~Shan and Weihong Deng. 2018.
\newblock Reliable crowdsourcing and deep locality-preserving learning for unconstrained facial expression recognition.
\newblock \emph{IEEE Transactions on Image Processing}, 28(1):356--370.

\bibitem[{Sharma and Thakur(2023)}]{sharma2023chatgpt}
Gaurav Sharma and Abhishek Thakur. 2023.
\newblock Chatgpt in drug discovery.

\bibitem[{Shinn et~al.(2023)Shinn, Cassano, Berman, Gopinath, Narasimhan, and Yao}]{shinn2023reflexion}
Noah Shinn, Federico Cassano, Edward Berman, Ashwin Gopinath, Karthik Narasimhan, and Shunyu Yao. 2023.
\newblock \href {http://arxiv.org/abs/2303.11366} {Reflexion: Language agents with verbal reinforcement learning}.

\bibitem[{Singh et~al.(2023)Singh, Blukis, Mousavian, Goyal, Xu, Tremblay, Fox, Thomason, and Garg}]{singh2023progprompt}
Ishika Singh, Valts Blukis, Arsalan Mousavian, Ankit Goyal, Danfei Xu, Jonathan Tremblay, Dieter Fox, Jesse Thomason, and Animesh Garg. 2023.
\newblock Progprompt: Generating situated robot task plans using large language models.
\newblock In \emph{2023 IEEE International Conference on Robotics and Automation (ICRA)}, pages 11523--11530. IEEE.

\bibitem[{Sinha and Khandait(2021)}]{sinha2021impact}
Ankur Sinha and Tanmay Khandait. 2021.
\newblock Impact of news on the commodity market: Dataset and results.
\newblock In \emph{Advances in Information and Communication: Proceedings of the 2021 Future of Information and Communication Conference (FICC), Volume 2}, pages 589--601. Springer.

\bibitem[{Song et~al.(2023{\natexlab{a}})Song, Wu, Washington, Sadler, Chao, and Su}]{song2023llmplanner}
Chan~Hee Song, Jiaman Wu, Clayton Washington, Brian~M. Sadler, Wei-Lun Chao, and Yu~Su. 2023{\natexlab{a}}.
\newblock Llm-planner: Few-shot grounded planning for embodied agents with large language models.
\newblock In \emph{Proceedings of the IEEE/CVF International Conference on Computer Vision (ICCV)}.

\bibitem[{Song et~al.(2023{\natexlab{b}})Song, Xiong, Zhu, Wu, Qian, Song, Huang, Li, Wang, Yao, Tian, and Li}]{song2023restgpt}
Yifan Song, Weimin Xiong, Dawei Zhu, Wenhao Wu, Han Qian, Mingbo Song, Hailiang Huang, Cheng Li, Ke~Wang, Rong Yao, Ye~Tian, and Sujian Li. 2023{\natexlab{b}}.
\newblock \href {http://arxiv.org/abs/2306.06624} {Restgpt: Connecting large language models with real-world restful apis}.

\bibitem[{Sorensen et~al.(2024)Sorensen, Jiang, Hwang, Levine, Pyatkin, West, Dziri, Lu, Rao, Bhagavatula et~al.}]{sorensen2024value}
Taylor Sorensen, Liwei Jiang, Jena~D Hwang, Sydney Levine, Valentina Pyatkin, Peter West, Nouha Dziri, Ximing Lu, Kavel Rao, Chandra Bhagavatula, et~al. 2024.
\newblock Value kaleidoscope: Engaging ai with pluralistic human values, rights, and duties.
\newblock In \emph{Proceedings of the AAAI Conference on Artificial Intelligence}, volume~38, pages 19937--19947.

\bibitem[{Srivastava et~al.(2022)Srivastava, Rastogi, Rao, Shoeb, Abid, Fisch, Brown, Santoro, Gupta, Garriga-Alonso et~al.}]{srivastava2022beyond}
Aarohi Srivastava, Abhinav Rastogi, Abhishek Rao, Abu Awal~Md Shoeb, Abubakar Abid, Adam Fisch, Adam~R Brown, Adam Santoro, Aditya Gupta, Adri{\`a} Garriga-Alonso, et~al. 2022.
\newblock Beyond the imitation game: Quantifying and extrapolating the capabilities of language models.
\newblock \emph{arXiv preprint arXiv:2206.04615}.

\bibitem[{Srivastava et~al.(2023)Srivastava, Rastogi, Rao, Shoeb, Abid, Fisch, Brown, Santoro, Gupta, Garriga-Alonso, Kluska, Lewkowycz, Agarwal, Power, Ray, Warstadt, Kocurek, Safaya, Tazarv, Xiang, Parrish, Nie, Hussain, Askell, Dsouza, Slone, Rahane, Iyer, Andreassen, Madotto, Santilli, Stuhlmüller, Dai, La, Lampinen, Zou, Jiang, Chen, Vuong, Gupta, Gottardi, Norelli, Venkatesh, Gholamidavoodi, Tabassum, Menezes, Kirubarajan, Mullokandov, Sabharwal, Herrick, Efrat, Erdem, Karakaş, Roberts, Loe, Zoph, Bojanowski, Özyurt, Hedayatnia, Neyshabur, Inden, Stein, Ekmekci, Lin, Howald, Orinion, Diao, Dour, Stinson, Argueta, Ramírez, Singh, Rathkopf, Meng, Baral, Wu, Callison-Burch, Waites, Voigt, Manning, Potts, Ramirez, Rivera, Siro, Raffel, Ashcraft, Garbacea, Sileo, Garrette, Hendrycks, Kilman, Roth, Freeman, Khashabi, Levy, González, Perszyk, Hernandez, Chen, Ippolito, Gilboa, Dohan, Drakard, Jurgens, Datta, Ganguli, Emelin, Kleyko, Yuret, Chen, Tam, Hupkes, Misra, Buzan, Mollo, Yang, Lee, Schrader,
  Shutova, Cubuk, Segal, Hagerman, Barnes, Donoway, Pavlick, Rodola, Lam, Chu, Tang, Erdem, Chang, Chi, Dyer, Jerzak, Kim, Manyasi, Zheltonozhskii, Xia, Siar, Martínez-Plumed, Happé, Chollet, Rong, Mishra, Winata, de~Melo, Kruszewski, Parascandolo, Mariani, Wang, Jaimovitch-López, Betz, Gur-Ari, Galijasevic, Kim, Rashkin, Hajishirzi, Mehta, Bogar, Shevlin, Schütze, Yakura, Zhang, Wong, Ng, Noble, Jumelet, Geissinger, Kernion, Hilton, Lee, Fisac, Simon, Koppel, Zheng, Zou, Kocoń, Thompson, Wingfield, Kaplan, Radom, Sohl-Dickstein, Phang, Wei, Yosinski, Novikova, Bosscher, Marsh, Kim, Taal, Engel, Alabi, Xu, Song, Tang, Waweru, Burden, Miller, Balis, Batchelder, Berant, Frohberg, Rozen, Hernandez-Orallo, Boudeman, Guerr, Jones, Tenenbaum, Rule, Chua, Kanclerz, Livescu, Krauth, Gopalakrishnan, Ignatyeva, Markert, Dhole, Gimpel, Omondi, Mathewson, Chiafullo, Shkaruta, Shridhar, McDonell, Richardson, Reynolds, Gao, Zhang, Dugan, Qin, Contreras-Ochando, Morency, Moschella, Lam, Noble, Schmidt, He, Colón,
  Metz, Şenel, Bosma, Sap, ter Hoeve, Farooqi, Faruqui, Mazeika, Baturan, Marelli, Maru, Quintana, Tolkiehn, Giulianelli, Lewis, Potthast, Leavitt, Hagen, Schubert, Baitemirova, Arnaud, McElrath, Yee, Cohen, Gu, Ivanitskiy, Starritt, Strube, Swędrowski, Bevilacqua, Yasunaga, Kale, Cain, Xu, Suzgun, Walker, Tiwari, Bansal, Aminnaseri, Geva, Gheini, T, Peng, Chi, Lee, Krakover, Cameron, Roberts, Doiron, Martinez, Nangia, Deckers, Muennighoff, Keskar, Iyer, Constant, Fiedel, Wen, Zhang, Agha, Elbaghdadi, Levy, Evans, Casares, Doshi, Fung, Liang, Vicol, Alipoormolabashi, Liao, Liang, Chang, Eckersley, Htut, Hwang, Miłkowski, Patil, Pezeshkpour, Oli, Mei, Lyu, Chen, Banjade, Rudolph, Gabriel, Habacker, Risco, Millière, Garg, Barnes, Saurous, Arakawa, Raymaekers, Frank, Sikand, Novak, Sitelew, LeBras, Liu, Jacobs, Zhang, Salakhutdinov, Chi, Lee, Stovall, Teehan, Yang, Singh, Mohammad, Anand, Dillavou, Shleifer, Wiseman, Gruetter, Bowman, Schoenholz, Han, Kwatra, Rous, Ghazarian, Ghosh, Casey, Bischoff,
  Gehrmann, Schuster, Sadeghi, Hamdan, Zhou, Srivastava, Shi, Singh, Asaadi, Gu, Pachchigar, Toshniwal, Upadhyay, Shyamolima, Debnath, Shakeri, Thormeyer, Melzi, Reddy, Makini, Lee, Torene, Hatwar, Dehaene, Divic, Ermon, Biderman, Lin, Prasad, Piantadosi, Shieber, Misherghi, Kiritchenko, Mishra, Linzen, Schuster, Li, Yu, Ali, Hashimoto, Wu, Desbordes, Rothschild, Phan, Wang, Nkinyili, Schick, Kornev, Tunduny, Gerstenberg, Chang, Neeraj, Khot, Shultz, Shaham, Misra, Demberg, Nyamai, Raunak, Ramasesh, Prabhu, Padmakumar, Srikumar, Fedus, Saunders, Zhang, Vossen, Ren, Tong, Zhao, Wu, Shen, Yaghoobzadeh, Lakretz, Song, Bahri, Choi, Yang, Hao, Chen, Belinkov, Hou, Hou, Bai, Seid, Zhao, Wang, Wang, Wang, and Wu}]{srivastava2023imitation}
Aarohi Srivastava, Abhinav Rastogi, Abhishek Rao, Abu Awal~Md Shoeb, Abubakar Abid, Adam Fisch, Adam~R. Brown, Adam Santoro, Aditya Gupta, Adrià Garriga-Alonso, Agnieszka Kluska, Aitor Lewkowycz, Akshat Agarwal, Alethea Power, Alex Ray, Alex Warstadt, Alexander~W. Kocurek, Ali Safaya, Ali Tazarv, Alice Xiang, Alicia Parrish, Allen Nie, Aman Hussain, Amanda Askell, Amanda Dsouza, Ambrose Slone, Ameet Rahane, Anantharaman~S. Iyer, Anders Andreassen, Andrea Madotto, Andrea Santilli, Andreas Stuhlmüller, Andrew Dai, Andrew La, Andrew Lampinen, Andy Zou, Angela Jiang, Angelica Chen, Anh Vuong, Animesh Gupta, Anna Gottardi, Antonio Norelli, Anu Venkatesh, Arash Gholamidavoodi, Arfa Tabassum, Arul Menezes, Arun Kirubarajan, Asher Mullokandov, Ashish Sabharwal, Austin Herrick, Avia Efrat, Aykut Erdem, Ayla Karakaş, B.~Ryan Roberts, Bao~Sheng Loe, Barret Zoph, Bartłomiej Bojanowski, Batuhan Özyurt, Behnam Hedayatnia, Behnam Neyshabur, Benjamin Inden, Benno Stein, Berk Ekmekci, Bill~Yuchen Lin, Blake Howald, Bryan
  Orinion, Cameron Diao, Cameron Dour, Catherine Stinson, Cedrick Argueta, César~Ferri Ramírez, Chandan Singh, Charles Rathkopf, Chenlin Meng, Chitta Baral, Chiyu Wu, Chris Callison-Burch, Chris Waites, Christian Voigt, Christopher~D. Manning, Christopher Potts, Cindy Ramirez, Clara~E. Rivera, Clemencia Siro, Colin Raffel, Courtney Ashcraft, Cristina Garbacea, Damien Sileo, Dan Garrette, Dan Hendrycks, Dan Kilman, Dan Roth, Daniel Freeman, Daniel Khashabi, Daniel Levy, Daniel~Moseguí González, Danielle Perszyk, Danny Hernandez, Danqi Chen, Daphne Ippolito, Dar Gilboa, David Dohan, David Drakard, David Jurgens, Debajyoti Datta, Deep Ganguli, Denis Emelin, Denis Kleyko, Deniz Yuret, Derek Chen, Derek Tam, Dieuwke Hupkes, Diganta Misra, Dilyar Buzan, Dimitri~Coelho Mollo, Diyi Yang, Dong-Ho Lee, Dylan Schrader, Ekaterina Shutova, Ekin~Dogus Cubuk, Elad Segal, Eleanor Hagerman, Elizabeth Barnes, Elizabeth Donoway, Ellie Pavlick, Emanuele Rodola, Emma Lam, Eric Chu, Eric Tang, Erkut Erdem, Ernie Chang,
  Ethan~A. Chi, Ethan Dyer, Ethan Jerzak, Ethan Kim, Eunice~Engefu Manyasi, Evgenii Zheltonozhskii, Fanyue Xia, Fatemeh Siar, Fernando Martínez-Plumed, Francesca Happé, Francois Chollet, Frieda Rong, Gaurav Mishra, Genta~Indra Winata, Gerard de~Melo, Germán Kruszewski, Giambattista Parascandolo, Giorgio Mariani, Gloria Wang, Gonzalo Jaimovitch-López, Gregor Betz, Guy Gur-Ari, Hana Galijasevic, Hannah Kim, Hannah Rashkin, Hannaneh Hajishirzi, Harsh Mehta, Hayden Bogar, Henry Shevlin, Hinrich Schütze, Hiromu Yakura, Hongming Zhang, Hugh~Mee Wong, Ian Ng, Isaac Noble, Jaap Jumelet, Jack Geissinger, Jackson Kernion, Jacob Hilton, Jaehoon Lee, Jaime~Fernández Fisac, James~B. Simon, James Koppel, James Zheng, James Zou, Jan Kocoń, Jana Thompson, Janelle Wingfield, Jared Kaplan, Jarema Radom, Jascha Sohl-Dickstein, Jason Phang, Jason Wei, Jason Yosinski, Jekaterina Novikova, Jelle Bosscher, Jennifer Marsh, Jeremy Kim, Jeroen Taal, Jesse Engel, Jesujoba Alabi, Jiacheng Xu, Jiaming Song, Jillian Tang, Joan
  Waweru, John Burden, John Miller, John~U. Balis, Jonathan Batchelder, Jonathan Berant, Jörg Frohberg, Jos Rozen, Jose Hernandez-Orallo, Joseph Boudeman, Joseph Guerr, Joseph Jones, Joshua~B. Tenenbaum, Joshua~S. Rule, Joyce Chua, Kamil Kanclerz, Karen Livescu, Karl Krauth, Karthik Gopalakrishnan, Katerina Ignatyeva, Katja Markert, Kaustubh~D. Dhole, Kevin Gimpel, Kevin Omondi, Kory Mathewson, Kristen Chiafullo, Ksenia Shkaruta, Kumar Shridhar, Kyle McDonell, Kyle Richardson, Laria Reynolds, Leo Gao, Li~Zhang, Liam Dugan, Lianhui Qin, Lidia Contreras-Ochando, Louis-Philippe Morency, Luca Moschella, Lucas Lam, Lucy Noble, Ludwig Schmidt, Luheng He, Luis~Oliveros Colón, Luke Metz, Lütfi~Kerem Şenel, Maarten Bosma, Maarten Sap, Maartje ter Hoeve, Maheen Farooqi, Manaal Faruqui, Mantas Mazeika, Marco Baturan, Marco Marelli, Marco Maru, Maria Jose~Ramírez Quintana, Marie Tolkiehn, Mario Giulianelli, Martha Lewis, Martin Potthast, Matthew~L. Leavitt, Matthias Hagen, Mátyás Schubert, Medina~Orduna
  Baitemirova, Melody Arnaud, Melvin McElrath, Michael~A. Yee, Michael Cohen, Michael Gu, Michael Ivanitskiy, Michael Starritt, Michael Strube, Michał Swędrowski, Michele Bevilacqua, Michihiro Yasunaga, Mihir Kale, Mike Cain, Mimee Xu, Mirac Suzgun, Mitch Walker, Mo~Tiwari, Mohit Bansal, Moin Aminnaseri, Mor Geva, Mozhdeh Gheini, Mukund~Varma T, Nanyun Peng, Nathan~A. Chi, Nayeon Lee, Neta Gur-Ari Krakover, Nicholas Cameron, Nicholas Roberts, Nick Doiron, Nicole Martinez, Nikita Nangia, Niklas Deckers, Niklas Muennighoff, Nitish~Shirish Keskar, Niveditha~S. Iyer, Noah Constant, Noah Fiedel, Nuan Wen, Oliver Zhang, Omar Agha, Omar Elbaghdadi, Omer Levy, Owain Evans, Pablo Antonio~Moreno Casares, Parth Doshi, Pascale Fung, Paul~Pu Liang, Paul Vicol, Pegah Alipoormolabashi, Peiyuan Liao, Percy Liang, Peter Chang, Peter Eckersley, Phu~Mon Htut, Pinyu Hwang, Piotr Miłkowski, Piyush Patil, Pouya Pezeshkpour, Priti Oli, Qiaozhu Mei, Qing Lyu, Qinlang Chen, Rabin Banjade, Rachel~Etta Rudolph, Raefer Gabriel, Rahel
  Habacker, Ramon Risco, Raphaël Millière, Rhythm Garg, Richard Barnes, Rif~A. Saurous, Riku Arakawa, Robbe Raymaekers, Robert Frank, Rohan Sikand, Roman Novak, Roman Sitelew, Ronan LeBras, Rosanne Liu, Rowan Jacobs, Rui Zhang, Ruslan Salakhutdinov, Ryan Chi, Ryan Lee, Ryan Stovall, Ryan Teehan, Rylan Yang, Sahib Singh, Saif~M. Mohammad, Sajant Anand, Sam Dillavou, Sam Shleifer, Sam Wiseman, Samuel Gruetter, Samuel~R. Bowman, Samuel~S. Schoenholz, Sanghyun Han, Sanjeev Kwatra, Sarah~A. Rous, Sarik Ghazarian, Sayan Ghosh, Sean Casey, Sebastian Bischoff, Sebastian Gehrmann, Sebastian Schuster, Sepideh Sadeghi, Shadi Hamdan, Sharon Zhou, Shashank Srivastava, Sherry Shi, Shikhar Singh, Shima Asaadi, Shixiang~Shane Gu, Shubh Pachchigar, Shubham Toshniwal, Shyam Upadhyay, Shyamolima, Debnath, Siamak Shakeri, Simon Thormeyer, Simone Melzi, Siva Reddy, Sneha~Priscilla Makini, Soo-Hwan Lee, Spencer Torene, Sriharsha Hatwar, Stanislas Dehaene, Stefan Divic, Stefano Ermon, Stella Biderman, Stephanie Lin, Stephen
  Prasad, Steven~T. Piantadosi, Stuart~M. Shieber, Summer Misherghi, Svetlana Kiritchenko, Swaroop Mishra, Tal Linzen, Tal Schuster, Tao Li, Tao Yu, Tariq Ali, Tatsu Hashimoto, Te-Lin Wu, Théo Desbordes, Theodore Rothschild, Thomas Phan, Tianle Wang, Tiberius Nkinyili, Timo Schick, Timofei Kornev, Titus Tunduny, Tobias Gerstenberg, Trenton Chang, Trishala Neeraj, Tushar Khot, Tyler Shultz, Uri Shaham, Vedant Misra, Vera Demberg, Victoria Nyamai, Vikas Raunak, Vinay Ramasesh, Vinay~Uday Prabhu, Vishakh Padmakumar, Vivek Srikumar, William Fedus, William Saunders, William Zhang, Wout Vossen, Xiang Ren, Xiaoyu Tong, Xinran Zhao, Xinyi Wu, Xudong Shen, Yadollah Yaghoobzadeh, Yair Lakretz, Yangqiu Song, Yasaman Bahri, Yejin Choi, Yichi Yang, Yiding Hao, Yifu Chen, Yonatan Belinkov, Yu~Hou, Yufang Hou, Yuntao Bai, Zachary Seid, Zhuoye Zhao, Zijian Wang, Zijie~J. Wang, Zirui Wang, and Ziyi Wu. 2023.
\newblock \href {http://arxiv.org/abs/2206.04615} {Beyond the imitation game: Quantifying and extrapolating the capabilities of language models}.

\bibitem[{Staab et~al.(2023)Staab, Vero, Balunovi{\'c}, and Vechev}]{staab2023beyond}
Robin Staab, Mark Vero, Mislav Balunovi{\'c}, and Martin Vechev. 2023.
\newblock Beyond memorization: Violating privacy via inference with large language models.
\newblock \emph{arXiv preprint arXiv:2310.07298}.

\bibitem[{Stolfo et~al.(2023)Stolfo, Jin, Shridhar, Schölkopf, and Sachan}]{stolfo2023causal}
Alessandro Stolfo, Zhijing Jin, Kumar Shridhar, Bernhard Schölkopf, and Mrinmaya Sachan. 2023.
\newblock \href {http://arxiv.org/abs/2210.12023} {A causal framework to quantify the robustness of mathematical reasoning with language models}.

\bibitem[{Sun et~al.(2023)Sun, Zhuang, Kong, Dai, and Zhang}]{NEURIPS2023_adaplanner}
Haotian Sun, Yuchen Zhuang, Lingkai Kong, Bo~Dai, and Chao Zhang. 2023.
\newblock \href {https://proceedings.neurips.cc/paper_files/paper/2023/file/b5c8c1c117618267944b2617add0a766-Paper-Conference.pdf} {Adaplanner: Adaptive planning from feedback with language models}.
\newblock In \emph{Advances in Neural Information Processing Systems}, volume~36, pages 58202--58245. Curran Associates, Inc.

\bibitem[{Sun et~al.(2024)Sun, Zheng, Xie, Liu, Chu, Qiu, Xu, Ding, Li, Geng, Wu, Wang, Chen, Yin, Ren, Fu, He, Yuan, Liu, Liu, Li, Dong, Cheng, Zhang, Heng, Dai, Luo, Wang, Wen, Qiu, Guo, Xiong, Liu, and Li}]{sun2024survey}
Jiankai Sun, Chuanyang Zheng, Enze Xie, Zhengying Liu, Ruihang Chu, Jianing Qiu, Jiaqi Xu, Mingyu Ding, Hongyang Li, Mengzhe Geng, Yue Wu, Wenhai Wang, Junsong Chen, Zhangyue Yin, Xiaozhe Ren, Jie Fu, Junxian He, Wu~Yuan, Qi~Liu, Xihui Liu, Yu~Li, Hao Dong, Yu~Cheng, Ming Zhang, Pheng~Ann Heng, Jifeng Dai, Ping Luo, Jingdong Wang, Ji-Rong Wen, Xipeng Qiu, Yike Guo, Hui Xiong, Qun Liu, and Zhenguo Li. 2024.
\newblock \href {http://arxiv.org/abs/2312.11562} {A survey of reasoning with foundation models}.

\bibitem[{Suzgun et~al.(2022)Suzgun, Scales, Sch{\"a}rli, Gehrmann, Tay, Chung, Chowdhery, Le, Chi, Zhou et~al.}]{suzgun2022challenging}
Mirac Suzgun, Nathan Scales, Nathanael Sch{\"a}rli, Sebastian Gehrmann, Yi~Tay, Hyung~Won Chung, Aakanksha Chowdhery, Quoc~V Le, Ed~H Chi, Denny Zhou, et~al. 2022.
\newblock Challenging big-bench tasks and whether chain-of-thought can solve them.
\newblock \emph{arXiv preprint arXiv:2210.09261}.

\bibitem[{Talmor et~al.(2018)Talmor, Herzig, Lourie, and Berant}]{talmor2018commonsenseqa}
Alon Talmor, Jonathan Herzig, Nicholas Lourie, and Jonathan Berant. 2018.
\newblock Commonsenseqa: A question answering challenge targeting commonsense knowledge.
\newblock \emph{arXiv preprint arXiv:1811.00937}.

\bibitem[{Tang et~al.(2024)Tang, Shalyminov, mei Wong, Burnsky, Vincent, Yang, Singh, Feng, Song, Su, Sun, Zhang, Mansour, and McKeown}]{tang2024tofueval}
Liyan Tang, Igor Shalyminov, Amy~Wing mei Wong, Jon Burnsky, Jake~W. Vincent, Yu'an Yang, Siffi Singh, Song Feng, Hwanjun Song, Hang Su, Lijia Sun, Yi~Zhang, Saab Mansour, and Kathleen McKeown. 2024.
\newblock \href {http://arxiv.org/abs/2402.13249} {Tofueval: Evaluating hallucinations of llms on topic-focused dialogue summarization}.

\bibitem[{Tang et~al.(2023)Tang, Deng, Lin, Han, Liang, Cao, and Sun}]{tang2023toolalpaca}
Qiaoyu Tang, Ziliang Deng, Hongyu Lin, Xianpei Han, Qiao Liang, Boxi Cao, and Le~Sun. 2023.
\newblock \href {http://arxiv.org/abs/2306.05301} {Toolalpaca: Generalized tool learning for language models with 3000 simulated cases}.

\bibitem[{Thirunavukarasu et~al.(2023)Thirunavukarasu, Ting, Elangovan, Gutierrez, Tan, and Ting}]{thirunavukarasu2023large}
Arun~James Thirunavukarasu, Darren Shu~Jeng Ting, Kabilan Elangovan, Laura Gutierrez, Ting~Fang Tan, and Daniel Shu~Wei Ting. 2023.
\newblock Large language models in medicine.
\newblock \emph{Nature medicine}, 29(8):1930--1940.

\bibitem[{Tobia(2020)}]{tobia2020testing}
Kevin~P Tobia. 2020.
\newblock Testing ordinary meaning.
\newblock \emph{Harv. L. Rev.}, 134:726.

\bibitem[{Turpin et~al.(2023)Turpin, Michael, Perez, and Bowman}]{turpin2023language}
Miles Turpin, Julian Michael, Ethan Perez, and Samuel~R. Bowman. 2023.
\newblock \href {http://arxiv.org/abs/2305.04388} {Language models don't always say what they think: Unfaithful explanations in chain-of-thought prompting}.

\bibitem[{Vaswani et~al.(2017)Vaswani, Shazeer, Parmar, Uszkoreit, Jones, Gomez, Kaiser, and Polosukhin}]{vaswani2017attention}
Ashish Vaswani, Noam Shazeer, Niki Parmar, Jakob Uszkoreit, Llion Jones, Aidan~N Gomez, {\L}ukasz Kaiser, and Illia Polosukhin. 2017.
\newblock Attention is all you need.
\newblock \emph{Advances in neural information processing systems}, 30.

\bibitem[{Vermetten et~al.(2022)Vermetten, van Stein, Caraffini, Minku, and Kononova}]{vermetten2022bias}
Diederick Vermetten, Bas van Stein, Fabio Caraffini, Leandro~L Minku, and Anna~V Kononova. 2022.
\newblock Bias: A toolbox for benchmarking structural bias in the continuous domain.
\newblock \emph{IEEE Transactions on Evolutionary Computation}, 26(6):1380--1393.

\bibitem[{Vidgen et~al.(2024)Vidgen, Agrawal, Ahmed, Akinwande, Al-Nuaimi, Alfaraj, Alhajjar, Aroyo, Bavalatti, Blili-Hamelin, Bollacker, Bomassani, Boston, Campos, Chakra, Chen, Coleman, Coudert, Derczynski, Dutta, Eisenberg, Ezick, Frase, Fuller, Gandikota, Gangavarapu, Gangavarapu, Gealy, Ghosh, Goel, Gohar, Goswami, Hale, Hutiri, Imperial, Jandial, Judd, Juefei-Xu, Khomh, Kailkhura, Kirk, Klyman, Knotz, Kuchnik, Kumar, Lengerich, Li, Liao, Long, Lu, Mai, Mammen, Manyeki, McGregor, Mehta, Mohammed, Moss, Nachman, Naganna, Nikanjam, Nushi, Oala, Orr, Parrish, Patlak, Pietri, Poursabzi-Sangdeh, Presani, Puletti, Röttger, Sahay, Santos, Scherrer, Sebag, Schramowski, Shahbazi, Sharma, Shen, Sistla, Tang, Testuggine, Thangarasa, Watkins, Weiss, Welty, Wilbers, Williams, Wu, Yadav, Yang, Zeng, Zhang, Zhdanov, Zhu, Liang, Mattson, and Vanschoren}]{vidgen2024introducing}
Bertie Vidgen, Adarsh Agrawal, Ahmed~M. Ahmed, Victor Akinwande, Namir Al-Nuaimi, Najla Alfaraj, Elie Alhajjar, Lora Aroyo, Trupti Bavalatti, Borhane Blili-Hamelin, Kurt Bollacker, Rishi Bomassani, Marisa~Ferrara Boston, Siméon Campos, Kal Chakra, Canyu Chen, Cody Coleman, Zacharie~Delpierre Coudert, Leon Derczynski, Debojyoti Dutta, Ian Eisenberg, James Ezick, Heather Frase, Brian Fuller, Ram Gandikota, Agasthya Gangavarapu, Ananya Gangavarapu, James Gealy, Rajat Ghosh, James Goel, Usman Gohar, Sujata Goswami, Scott~A. Hale, Wiebke Hutiri, Joseph~Marvin Imperial, Surgan Jandial, Nick Judd, Felix Juefei-Xu, Foutse Khomh, Bhavya Kailkhura, Hannah~Rose Kirk, Kevin Klyman, Chris Knotz, Michael Kuchnik, Shachi~H. Kumar, Chris Lengerich, Bo~Li, Zeyi Liao, Eileen~Peters Long, Victor Lu, Yifan Mai, Priyanka~Mary Mammen, Kelvin Manyeki, Sean McGregor, Virendra Mehta, Shafee Mohammed, Emanuel Moss, Lama Nachman, Dinesh~Jinenhally Naganna, Amin Nikanjam, Besmira Nushi, Luis Oala, Iftach Orr, Alicia Parrish, Cigdem
  Patlak, William Pietri, Forough Poursabzi-Sangdeh, Eleonora Presani, Fabrizio Puletti, Paul Röttger, Saurav Sahay, Tim Santos, Nino Scherrer, Alice~Schoenauer Sebag, Patrick Schramowski, Abolfazl Shahbazi, Vin Sharma, Xudong Shen, Vamsi Sistla, Leonard Tang, Davide Testuggine, Vithursan Thangarasa, Elizabeth~Anne Watkins, Rebecca Weiss, Chris Welty, Tyler Wilbers, Adina Williams, Carole-Jean Wu, Poonam Yadav, Xianjun Yang, Yi~Zeng, Wenhui Zhang, Fedor Zhdanov, Jiacheng Zhu, Percy Liang, Peter Mattson, and Joaquin Vanschoren. 2024.
\newblock \href {http://arxiv.org/abs/2404.12241} {Introducing v0.5 of the ai safety benchmark from mlcommons}.

\bibitem[{Wang et~al.(2024{\natexlab{a}})Wang, Wang, Xu, Zhang, Gu, Jia, Wang, Xu, Yan, Zhang, and Sang}]{wang2024amber}
Junyang Wang, Yuhang Wang, Guohai Xu, Jing Zhang, Yukai Gu, Haitao Jia, Jiaqi Wang, Haiyang Xu, Ming Yan, Ji~Zhang, and Jitao Sang. 2024{\natexlab{a}}.
\newblock \href {http://arxiv.org/abs/2311.07397} {Amber: An llm-free multi-dimensional benchmark for mllms hallucination evaluation}.

\bibitem[{Wang et~al.(2024{\natexlab{b}})Wang, Ma, Feng, Zhang, Yang, Zhang, Chen, Tang, Chen, Lin, Zhao, Wei, and Wen}]{Wang_2024}
Lei Wang, Chen Ma, Xueyang Feng, Zeyu Zhang, Hao Yang, Jingsen Zhang, Zhiyuan Chen, Jiakai Tang, Xu~Chen, Yankai Lin, Wayne~Xin Zhao, Zhewei Wei, and Jirong Wen. 2024{\natexlab{b}}.
\newblock \href {https://doi.org/10.1007/s11704-024-40231-1} {A survey on large language model based autonomous agents}.
\newblock \emph{Frontiers of Computer Science}, 18(6).

\bibitem[{Wang et~al.(2024{\natexlab{c}})Wang, Long, Fan, Wei, and Huang}]{wang2024benchmark}
Siyuan Wang, Zhuohan Long, Zhihao Fan, Zhongyu Wei, and Xuanjing Huang. 2024{\natexlab{c}}.
\newblock Benchmark self-evolving: A multi-agent framework for dynamic llm evaluation.
\newblock \emph{arXiv preprint arXiv:2402.11443}.

\bibitem[{Wang et~al.(2018)Wang, Durrett, and Erk}]{wang2018modeling}
Su~Wang, Greg Durrett, and Katrin Erk. 2018.
\newblock Modeling semantic plausibility by injecting world knowledge.
\newblock \emph{arXiv preprint arXiv:1804.00619}.

\bibitem[{Wang et~al.(2023{\natexlab{a}})Wang, Kordi, Mishra, Liu, Smith, Khashabi, and Hajishirzi}]{wang2023selfinstruct}
Yizhong Wang, Yeganeh Kordi, Swaroop Mishra, Alisa Liu, Noah~A. Smith, Daniel Khashabi, and Hannaneh Hajishirzi. 2023{\natexlab{a}}.
\newblock \href {http://arxiv.org/abs/2212.10560} {Self-instruct: Aligning language models with self-generated instructions}.

\bibitem[{Wang et~al.(2023{\natexlab{b}})Wang, Cai, Chen, Liu, Ma, and Liang}]{wang2023describe}
Zihao Wang, Shaofei Cai, Guanzhou Chen, Anji Liu, Xiaojian Ma, and Yitao Liang. 2023{\natexlab{b}}.
\newblock Describe, explain, plan and select: Interactive planning with large language models enables open-world multi-task agents.
\newblock \emph{arXiv preprint arXiv:2302.01560}.

\bibitem[{Wei et~al.(2024)Wei, Yao, Ton, Guo, Estornell, and Liu}]{wei2024measuring}
Jiaheng Wei, Yuanshun Yao, Jean-Francois Ton, Hongyi Guo, Andrew Estornell, and Yang Liu. 2024.
\newblock Measuring and reducing llm hallucination without gold-standard answers via expertise-weighting.
\newblock \emph{arXiv preprint arXiv:2402.10412}.

\bibitem[{Wessel et~al.(2023)Wessel, Horych, Ruas, Aizawa, Gipp, and Spinde}]{wessel2023introducing}
Martin Wessel, Tom{\'a}s Horych, Terry Ruas, Akiko Aizawa, Bela Gipp, and Timo Spinde. 2023.
\newblock Introducing mbib-the first media bias identification benchmark task and dataset collection.
\newblock In \emph{Proceedings of the 46th International ACM SIGIR Conference on Research and Development in Information Retrieval}, pages 2765--2774.

\bibitem[{Weston et~al.(2015)Weston, Bordes, Chopra, Rush, Van~Merri{\"e}nboer, Joulin, and Mikolov}]{weston2015towards}
Jason Weston, Antoine Bordes, Sumit Chopra, Alexander~M Rush, Bart Van~Merri{\"e}nboer, Armand Joulin, and Tomas Mikolov. 2015.
\newblock Towards ai-complete question answering: A set of prerequisite toy tasks.
\newblock \emph{arXiv preprint arXiv:1502.05698}.

\bibitem[{Wu and Aji(2023)}]{wu2023style}
Minghao Wu and Alham~Fikri Aji. 2023.
\newblock \href {http://arxiv.org/abs/2307.03025} {Style over substance: Evaluation biases for large language models}.

\bibitem[{Wu et~al.(2023)Wu, Irsoy, Lu, Dabravolski, Dredze, Gehrmann, Kambadur, Rosenberg, and Mann}]{wu2023bloomberggpt}
Shijie Wu, Ozan Irsoy, Steven Lu, Vadim Dabravolski, Mark Dredze, Sebastian Gehrmann, Prabhanjan Kambadur, David Rosenberg, and Gideon Mann. 2023.
\newblock Bloomberggpt: A large language model for finance.
\newblock \emph{arXiv preprint arXiv:2303.17564}.

\bibitem[{Xie et~al.(2023)Xie, Han, Zhang, Lai, Peng, Lopez-Lira, and Huang}]{xie2023pixiu}
Qianqian Xie, Weiguang Han, Xiao Zhang, Yanzhao Lai, Min Peng, Alejandro Lopez-Lira, and Jimin Huang. 2023.
\newblock Pixiu: A large language model, instruction data and evaluation benchmark for finance.
\newblock \emph{arXiv preprint arXiv:2306.05443}.

\bibitem[{Xu et~al.(2023{\natexlab{a}})Xu, Lin, Han, Zhao, Liu, and Cambria}]{xu2023large}
Fangzhi Xu, Qika Lin, Jiawei Han, Tianzhe Zhao, Jun Liu, and Erik Cambria. 2023{\natexlab{a}}.
\newblock \href {http://arxiv.org/abs/2306.09841} {Are large language models really good logical reasoners? a comprehensive evaluation and beyond}.

\bibitem[{Xu et~al.(2023{\natexlab{b}})Xu, Liu, Yan, Xu, Si, Zhou, Yi, Gao, Sang, Zhang, Zhang, Peng, Huang, and Zhou}]{xu2023cvalues}
Guohai Xu, Jiayi Liu, Ming Yan, Haotian Xu, Jinghui Si, Zhuoran Zhou, Peng Yi, Xing Gao, Jitao Sang, Rong Zhang, Ji~Zhang, Chao Peng, Fei Huang, and Jingren Zhou. 2023{\natexlab{b}}.
\newblock \href {http://arxiv.org/abs/2307.09705} {Cvalues: Measuring the values of chinese large language models from safety to responsibility}.

\bibitem[{Xu et~al.(2023{\natexlab{c}})Xu, Hong, Li, Hu, Chen, and Zhang}]{xu2023tool}
Qiantong Xu, Fenglu Hong, Bo~Li, Changran Hu, Zhengyu Chen, and Jian Zhang. 2023{\natexlab{c}}.
\newblock \href {http://arxiv.org/abs/2305.16504} {On the tool manipulation capability of open-source large language models}.

\bibitem[{Yan et~al.(2024{\natexlab{a}})Yan, Li, Xu, Dong, Zhang, Ren, and Cheng}]{yan2024protecting}
Biwei Yan, Kun Li, Minghui Xu, Yueyan Dong, Yue Zhang, Zhaochun Ren, and Xiuzhen Cheng. 2024{\natexlab{a}}.
\newblock \href {http://arxiv.org/abs/2403.05156} {On protecting the data privacy of large language models (llms): A survey}.

\bibitem[{Yan et~al.(2024{\natexlab{b}})Yan, Mao, Ji, Zhang, Patil, Stoica, and Gonzalez}]{berkeley-function-calling-leaderboard}
Fanjia Yan, Huanzhi Mao, Charlie Cheng-Jie Ji, Tianjun Zhang, Shishir~G. Patil, Ion Stoica, and Joseph~E. Gonzalez. 2024{\natexlab{b}}.
\newblock Berkeley function calling leaderboard.

\bibitem[{Yan et~al.(2014)Yan, Li, Wang, Zhao, Liu, Chen, and Fu}]{yan2014casme}
Wen-Jing Yan, Xiaobai Li, Su-Jing Wang, Guoying Zhao, Yong-Jin Liu, Yu-Hsin Chen, and Xiaolan Fu. 2014.
\newblock Casme ii: An improved spontaneous micro-expression database and the baseline evaluation.
\newblock \emph{PloS one}, 9(1):e86041.

\bibitem[{Yang et~al.(2023)Yang, Sun, and Wan}]{yang2023new}
Shiping Yang, Renliang Sun, and Xiaojun Wan. 2023.
\newblock \href {http://arxiv.org/abs/2310.06498} {A new benchmark and reverse validation method for passage-level hallucination detection}.

\bibitem[{Yang et~al.(2018)Yang, Qi, Zhang, Bengio, Cohen, Salakhutdinov, and Manning}]{yang2018hotpotqa}
Zhilin Yang, Peng Qi, Saizheng Zhang, Yoshua Bengio, William~W Cohen, Ruslan Salakhutdinov, and Christopher~D Manning. 2018.
\newblock Hotpotqa: A dataset for diverse, explainable multi-hop question answering.
\newblock \emph{arXiv preprint arXiv:1809.09600}.

\bibitem[{Yao et~al.(2023{\natexlab{a}})Yao, Chen, Yang, and Narasimhan}]{yao2023webshop}
Shunyu Yao, Howard Chen, John Yang, and Karthik Narasimhan. 2023{\natexlab{a}}.
\newblock \href {http://arxiv.org/abs/2207.01206} {Webshop: Towards scalable real-world web interaction with grounded language agents}.

\bibitem[{Yao et~al.(2023{\natexlab{b}})Yao, Zhao, Yu, Du, Shafran, Narasimhan, and Cao}]{yao2023react}
Shunyu Yao, Jeffrey Zhao, Dian Yu, Nan Du, Izhak Shafran, Karthik Narasimhan, and Yuan Cao. 2023{\natexlab{b}}.
\newblock \href {http://arxiv.org/abs/2210.03629} {React: Synergizing reasoning and acting in language models}.

\bibitem[{Yao et~al.(2024)Yao, Duan, Xu, Cai, Sun, and Zhang}]{Yao_2024}
Yifan Yao, Jinhao Duan, Kaidi Xu, Yuanfang Cai, Zhibo Sun, and Yue Zhang. 2024.
\newblock \href {https://doi.org/10.1016/j.hcc.2024.100211} {A survey on large language model (llm) security and privacy: The good, the bad, and the ugly}.
\newblock \emph{High-Confidence Computing}, page 100211.

\bibitem[{Yip et~al.(2024)Yip, Esmradi, and Chan}]{yip2024novel}
Daniel~Wankit Yip, Aysan Esmradi, and Chun~Fai Chan. 2024.
\newblock \href {http://arxiv.org/abs/2401.00991} {A novel evaluation framework for assessing resilience against prompt injection attacks in large language models}.

\bibitem[{Young et~al.(2022)Young, Bao, Bensemann, and Witbrock}]{young2022abductionrules}
Nathan Young, Qiming Bao, Joshua Bensemann, and Michael Witbrock. 2022.
\newblock Abductionrules: Training transformers to explain unexpected inputs.
\newblock \emph{arXiv preprint arXiv:2203.12186}.

\bibitem[{Yu et~al.(2023)Yu, Gileadi, Fu, Kirmani, Lee, Gonzalez~Arenas, Lewis~Chiang, Erez, Hasenclever, Humplik, Ichter, Xiao, Xu, Zeng, Zhang, Heess, Sadigh, Tan, Tassa, and Xia}]{yu2023lang2reward}
Wenhao Yu, Nimrod Gileadi, Chuyuan Fu, Sean Kirmani, Kuang-Huei Lee, Montse Gonzalez~Arenas, Hao-Tien Lewis~Chiang, Tom Erez, Leonard Hasenclever, Jan Humplik, Brian Ichter, Ted Xiao, Peng Xu, Andy Zeng, Tingnan Zhang, Nicolas Heess, Dorsa Sadigh, Jie Tan, Yuval Tassa, and Fei Xia. 2023.
\newblock Language to rewards for robotic skill synthesis.
\newblock \emph{Arxiv preprint arXiv:2306.08647}.

\bibitem[{Yuan et~al.(2023)Yuan, Yuan, Tan, Wang, and Huang}]{yuan2023large}
Zheng Yuan, Hongyi Yuan, Chuanqi Tan, Wei Wang, and Songfang Huang. 2023.
\newblock \href {http://arxiv.org/abs/2304.02015} {How well do large language models perform in arithmetic tasks?}

\bibitem[{Yuan et~al.(2024)Yuan, Xiong, Zeng, Yu, Jia, Song, and Li}]{yuan2024rigorllm}
Zhuowen Yuan, Zidi Xiong, Yi~Zeng, Ning Yu, Ruoxi Jia, Dawn Song, and Bo~Li. 2024.
\newblock \href {http://arxiv.org/abs/2403.13031} {Rigorllm: Resilient guardrails for large language models against undesired content}.

\bibitem[{Zhan et~al.(2024)Zhan, Liang, Ying, and Kang}]{zhan2024injecagent}
Qiusi Zhan, Zhixiang Liang, Zifan Ying, and Daniel Kang. 2024.
\newblock Injecagent: Benchmarking indirect prompt injections in tool-integrated large language model agents.
\newblock \emph{arXiv preprint arXiv:2403.02691}.

\bibitem[{Zhang et~al.(2023)Zhang, Bao, Zhang, Wang, Feng, and He}]{Zhang_2023}
Jizhi Zhang, Keqin Bao, Yang Zhang, Wenjie Wang, Fuli Feng, and Xiangnan He. 2023.
\newblock \href {https://doi.org/10.1145/3604915.3608860} {Is chatgpt fair for recommendation? evaluating fairness in large language model recommendation}.
\newblock In \emph{Proceedings of the 17th ACM Conference on Recommender Systems}, RecSys ’23. ACM.

\bibitem[{Zhang et~al.(2024{\natexlab{a}})Zhang, Li, Li, Shi, and Jin}]{zhang2024codeagent}
Kechi Zhang, Jia Li, Ge~Li, Xianjie Shi, and Zhi Jin. 2024{\natexlab{a}}.
\newblock \href {http://arxiv.org/abs/2401.07339} {Codeagent: Enhancing code generation with tool-integrated agent systems for real-world repo-level coding challenges}.

\bibitem[{Zhang et~al.(2024{\natexlab{b}})Zhang, Peng, Tian, Zhou, Jin, Song, Mi, and Meng}]{zhang2024self}
Xiaoying Zhang, Baolin Peng, Ye~Tian, Jingyan Zhou, Lifeng Jin, Linfeng Song, Haitao Mi, and Helen Meng. 2024{\natexlab{b}}.
\newblock Self-alignment for factuality: Mitigating hallucinations in llms via self-evaluation.
\newblock \emph{arXiv preprint arXiv:2402.09267}.

\bibitem[{Zhang and Yang(2023)}]{zhang2023xuanyuan}
Xuanyu Zhang and Qing Yang. 2023.
\newblock Xuanyuan 2.0: A large chinese financial chat model with hundreds of billions parameters.
\newblock In \emph{Proceedings of the 32nd ACM International Conference on Information and Knowledge Management}, pages 4435--4439.

\bibitem[{Zhang et~al.(2018)Zhang, Dai, Kozareva, Smola, and Song}]{zhang2017variational}
Yuyu Zhang, Hanjun Dai, Zornitsa Kozareva, Alexander~J Smola, and Le~Song. 2018.
\newblock Variational reasoning for question answering with knowledge graph.
\newblock In \emph{AAAI}.

\bibitem[{Zhao et~al.(2023)Zhao, Li, Li, and Pietik{\"a}inen}]{zhao2023facial}
Guoying Zhao, Xiaobai Li, Yante Li, and Matti Pietik{\"a}inen. 2023.
\newblock Facial micro-expressions: an overview.
\newblock \emph{Proceedings of the IEEE}.

\bibitem[{Zheng et~al.(2023{\natexlab{a}})Zheng, Huang, Zhao, Zhong, and Wang}]{zheng2023NaviLLM}
Duo Zheng, Shijia Huang, Lin Zhao, Yiwu Zhong, and Liwei Wang. 2023{\natexlab{a}}.
\newblock \href {http://arxiv.org/abs/2312.02010} {Towards learning a generalist model for embodied navigation}.

\bibitem[{Zheng et~al.(2024)Zheng, Chiang, Sheng, Zhuang, Wu, Zhuang, Lin, Li, Li, Xing et~al.}]{zheng2024judging}
Lianmin Zheng, Wei-Lin Chiang, Ying Sheng, Siyuan Zhuang, Zhanghao Wu, Yonghao Zhuang, Zi~Lin, Zhuohan Li, Dacheng Li, Eric Xing, et~al. 2024.
\newblock Judging llm-as-a-judge with mt-bench and chatbot arena.
\newblock \emph{Advances in Neural Information Processing Systems}, 36.

\bibitem[{Zheng et~al.(2023{\natexlab{b}})Zheng, Huang, and Chang}]{zheng2023does}
Shen Zheng, Jie Huang, and Kevin Chen-Chuan Chang. 2023{\natexlab{b}}.
\newblock \href {http://arxiv.org/abs/2304.10513} {Why does chatgpt fall short in providing truthful answers?}

\bibitem[{Zhou et~al.(2023{\natexlab{a}})Zhou, Hong, and Wu}]{zhou2023navgpt}
Gengze Zhou, Yicong Hong, and Qi~Wu. 2023{\natexlab{a}}.
\newblock Navgpt: Explicit reasoning in vision-and-language navigation with large language models.
\newblock \emph{arXiv preprint arXiv:2305.16986}.

\bibitem[{Zhou et~al.(2023{\natexlab{b}})Zhou, Xu, Zhu, Zhou, Lo, Sridhar, Cheng, Ou, Bisk, Fried, Alon, and Neubig}]{zhou2023webarena}
Shuyan Zhou, Frank~F. Xu, Hao Zhu, Xuhui Zhou, Robert Lo, Abishek Sridhar, Xianyi Cheng, Tianyue Ou, Yonatan Bisk, Daniel Fried, Uri Alon, and Graham Neubig. 2023{\natexlab{b}}.
\newblock \href {http://arxiv.org/abs/2307.13854} {Webarena: A realistic web environment for building autonomous agents}.

\bibitem[{Zielinski et~al.(2023)Zielinski, Winker, Aggarwal, Ferris, Heinemann, Lape{\~n}a~Jr, Pai, Ing, Citrome et~al.}]{zielinski2023wame}
Chris Zielinski, Margaret Winker, Rakesh Aggarwal, Lorraine Ferris, Markus Heinemann, Jose~Florencio Lape{\~n}a~Jr, Sanjay Pai, Edsel Ing, Leslie Citrome, et~al. 2023.
\newblock Wame recommendations on chatgpt and chatbots in relation to scholarly publications.

\end{thebibliography}

\end{document}